\documentclass[nohyperref]{article}

\usepackage{microtype}
\usepackage{graphicx}
\usepackage{subfigure}
\usepackage{booktabs} %
\usepackage[dvipsnames,table]{xcolor}

\usepackage{amsmath}

\DeclareMathOperator*{\argmin}{arg\,min}

\newcommand{\fig}[1]{Figure~\ref{#1}}
\newcommand{\sect}[1]{Section~\ref{#1}}
\newcommand{\tbl}[1]{Table~\ref{#1}}
\newcommand{\eqn}[1]{Eqn.~\ref{#1}}
\newcommand{\app}[1]{Appendix~\ref{#1}}

\newcommand{\ignorethis}[1]{}

\newcommand{\eg}{{\em e.g.\,}}

\usepackage{xcolor}
\usepackage{amsthm}
\usepackage{framed}

\newcommand{\xpar}[1]{$\triangleright$ \textbf{#1} \\}

\newcommand{\comm}[1]{}

\newcommand{\red}[1]{\textcolor{red}{#1}}
\newcommand{\blue}[1]{\textcolor{blue}{#1}}

\newcommand{\cL}{\mathcal{L}}

\newcommand\blfootnote[1]{%
  \begingroup
  \renewcommand\thefootnote{}\footnote{#1}%
  \addtocounter{footnote}{-1}%
  \endgroup
}
\newcommand{\bc}{\mathbf{c}}
\newcommand{\bx}{\mathbf{x}}
\newcommand{\by}{\mathbf{y}}
\newcommand{\bz}{\mathbf{z}}

\makeatletter
\def\blfootnote{\gdef\@thefnmark{}\@footnotetext}
\makeatother

\usepackage[accepted]{icml2023}

\usepackage{amsmath}
\usepackage{amssymb}
\usepackage{mathtools}
\usepackage{amsthm}
\usepackage{bbding}

\usepackage[colorlinks=true,linkcolor=blue,citecolor=teal]{hyperref}
\usepackage[capitalize,noabbrev]{cleveref}

\theoremstyle{plain}

\theoremstyle{definition}

\theoremstyle{remark}

\usepackage[textsize=tiny]{todonotes}

\icmltitlerunning{Straightening Out the Straight-Through Estimator}

\usepackage[utf8]{inputenc} %
\usepackage[T1]{fontenc}    %
\usepackage{nicefrac}       %
\usepackage{microtype}      %
\usepackage{xcolor}         %
\usepackage{lipsum}  
\usepackage[makeroom]{cancel}
\usepackage{mathtools}
\usepackage{caption}
\usepackage{multirow}

\usepackage{enumitem}

\begin{document}

\twocolumn[
\icmltitle{Straightening Out the Straight-Through Estimator:\\
Overcoming Optimization Challenges in Vector Quantized Networks}

\icmlsetsymbol{equal}{*}

\begin{icmlauthorlist}
\icmlauthor{Minyoung Huh}{csail}
\icmlauthor{Brian Cheung}{csail,bcs}
\icmlauthor{Pulkit Agrawal}{csail}
\icmlauthor{Phillip Isola}{csail}
\end{icmlauthorlist}

\icmlaffiliation{csail}{MIT CSAIL}
\icmlaffiliation{bcs}{MIT BCS}

\icmlcorrespondingauthor{Minyoung Huh}{minhuh@mit.edu}

\vskip 0.3in
]

\printAffiliationsAndNotice{\icmlEqualContribution} %

\begin{abstract}
This work examines the challenges of training neural networks using vector quantization using straight-through estimation. We find that a primary cause of training instability is the discrepancy between the model embedding and the code-vector distribution. We identify the factors that contribute to this issue, including the codebook gradient sparsity and the asymmetric nature of the commitment loss, which leads to misaligned code-vector assignments. We propose to address this issue via affine re-parameterization of the code vectors. Additionally, we introduce an alternating optimization to reduce the gradient error introduced by the straight-through estimation. Moreover, we propose an improvement to the commitment loss to ensure better alignment between the codebook representation and the model embedding. These optimization methods improve the mathematical approximation of the straight-through estimation and, ultimately, the model performance. We demonstrate the effectiveness of our methods on several common model architectures, such as AlexNet, ResNet, and ViT, across various tasks, including image classification and generative modeling. \\ 
{\fontsize{8.5pt}{8.5pt}\selectfont
Project\;page:\;\url{minyoungg.github.io/vqtorch}
\vspace{-0.2in}
}
\end{abstract}

\begin{figure}[t!]
     \centering
     \includegraphics[width=0.9\linewidth]{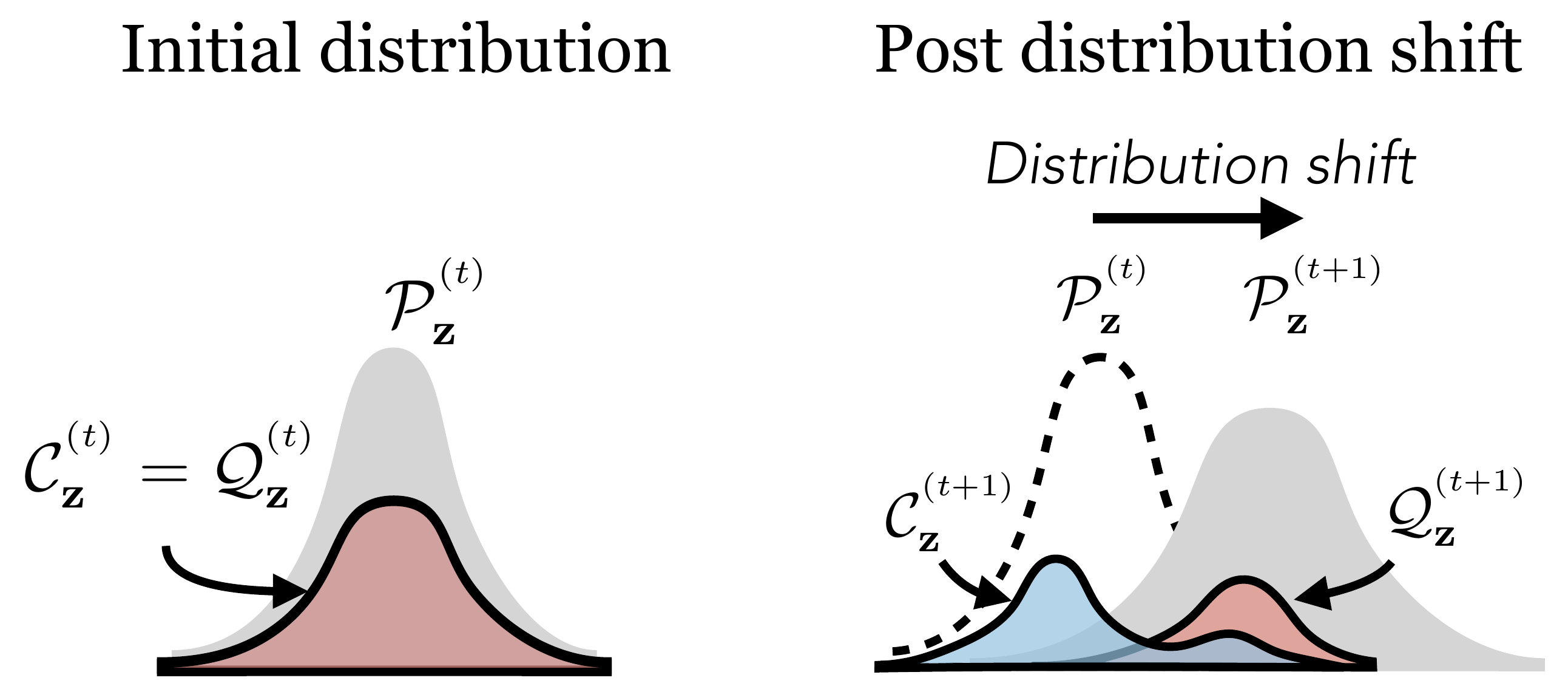}
     \caption{\small\textbf{Illustration of internal codebook covariate shift:} 
     During training, the embedding distribution $\mathcal{P}_{\bz}$ drifts from initialization. When the model undergoes a distributional shift, the codebook $\mathcal{C}_{\bz}$~(\textcolor{CornflowerBlue}{blue}) is misaligned with $\mathcal{P}_{\bz}$. The code-vectors that have assignments are denoted with $\mathcal{Q}_{\bz}$~(\textcolor{RedOrange}{red}) and initialized to overlap with $\mathcal{C}_{\bz}$. With training, $\mathcal{Q}_{\bz}$ diverge and are misaligned with $\mathcal{P}_{\bz}$. Code-vectors without assignment do not receive gradients and are no longer trained, which leads to bifurcation in the codebook distribution and ultimately leads to index collapse. 
     }
     \label{fig:drift}
     \vspace{-0.1in}
\end{figure}

\section{Introduction and related works}
Vector-quantization~\cite{gray1984vector} enables deep neural networks to learn discrete representations by quantizing features into clusters 
referred to as ``code-vectors'' or ``codes''. Vector-Quantization (VQ) is a parametric online K-means algorithm~\cite{caron2018deep} that has an explicit bias for compression and competition, which serves as a good prior for learning disentangled features for downstream tasks. 
To name a few, VQs have shown impressive results on image generation~\cite{van2017neural,ramesh2021zero,esser2020taming,chang2023muse}, image representation learning~\cite{caron2020unsupervised}, speech generation~\cite{dhariwal2020jukebox}, speech representation learning~\cite{chung2020vector}, and even decision-making~\cite{ozair2021vector}. 
While powerful, vector-quantized networks (VQNs) are notoriously difficult to optimize and require esoteric knowledge to efficiently train them. Hence, constructing algorithms to improve training stability has been a topic of great interest. 

First introduced in context of generative models~\cite{van2017neural}, the vector-quantization layer in VQNs maps the continuous embedding (or feature representation) $\mathcal{P}_{\bz}$ into a discrete embedding $\mathcal{Q}_{\bz}$ using the codebook $\mathcal{C}_{\bz}$. 
As the discretization function is not continuously differentiable, a widely used technique to optimize VQNs is via straight-through estimation. This bypasses the non-differentiable discretization function~\cite{bengio2013estimating}, allowing it to be optimizable by standard deep learning libraries~\cite{NEURIPS2019_9015}. Of course, straight-through estimating a selection function has negative ramifications on training. 
Among many training challenges, the most well-documented is model collapse or``index collapse'' wherein only a small fraction of codes are used during training.
While the root cause of the collapse is not well understood, there have been abundant efforts to mitigate index collapse: EMA~\cite{van2017neural}, codebook reset~\cite{lancucki2020robust,zeghidour2021soundstream,dhariwal2020jukebox},probabilistic/stochastic re-formulation~\cite{roy2018theory,takida2022sq}, equipartition assumption using optimal transport~\cite{asano2019self}, and many more (See~\sect{sec:sec3}). 
While many of these methods directly reduce the extent of index collapse, to the best of our knowledge, there has not been any work that investigates the root cause of the instability in the first place. Hence, our work aims to systematically investigate the cause of the model collapse and provide methods to address the common pitfalls that stem from unstable optimization.
Concretely, our contributions are as follows:
\begin{itemize}[noitemsep,topsep=0pt]
  \item We provide new insights into understanding VQ networks by formulating commitment loss as a divergence measure. Doing so allows us to understand better why the divergence occurs.
  \item To reduce this divergence, we propose an affine re-parameterization of the code-vectors that can better match the moments of the embedding representation. This alone drastically reduces the model collapse.
  \item Lastly, We provide a set of improvements on the existing optimization techniques, such as alternating optimization and synchronized commitment loss. Both these methods are simple and more mathematically correct update rules that result in improvements over the standard approach.
\end{itemize}

\section{Preliminaries}
We denote $x$ as a scalar, $\bx$ as a vector, $X$ as a matrix, $\mathcal{X}$ as a distribution or a set, $f(\cdot)$ as a function, $F(\cdot)$ as a composition of functions, and $\mathcal{L}(\cdot)$ for loss function. 

\xpar{Deep neural networks}
A feed-forward neural network is defined as a composition of parametric linear functions $f_i$ (\eg fully-connected, convolutional layer) and non-linearities $\sigma$ (\eg ReLU):

\vspace{-0.15in}
{\small
\begin{align}
\hat \by &= \underbracket[0.140ex]{f_n \circ \sigma \circ \cdots \circ \sigma \circ f_{i+1}}_{G} 
      \circ
      \underbracket[0.140ex]{f_{i} \circ \sigma \circ \cdots \circ f_2 \circ \sigma \circ f_1}_{F}
      \circ \; \bx \\
      &= G ( F ( \bx ) )
\end{align}
}
\vspace{-0.15in}

In the context of generative modeling, $F(\cdot)$ is referred to as the encoder and $G(\cdot)$ as the decoder. The network is trained by minimizing the empirical risk $\mathcal{L}_{\mathsf{task}}(\cdot)$ with dataset $\mathcal{D}$:

\vspace{-0.15in}
\begin{align}
\min_{F, G} \mathbb{E}_{(\bx, \by) \sim \mathcal{D}} \left[ \mathcal{L}_{\mathsf{task}} \left( G(F(\bx)), \by \right) \right] 
\end{align}
\vspace{-0.15in}

\xpar{Vector-quantized networks}
A vector-quantized network~(VQN) is a neural-network consisting of a vector-quantization layer $h(\cdot, \cdot)$:

\vspace{-0.15in}
\begin{align}
\hat \by = G( h (  F( \bx ), C )) = G( h (  \bz_e, C )) = G(\bz_q)
\end{align}
\vspace{-0.15in}

The VQ layer $h(\cdot)$ quantizes the embedding $\bz_e = F(\bx)$ by selecting a vector from a collection of $m$ vectors. The individual vector $\bc_i$ is referred to as the code-vector, the index $i$ as the code, and the collection of the code-vectors as the codebook $\mathcal{C} = \{ \bc_1, \bc_2, \dots \bc_m\}$. Here on out, we omit writing the codebook $C$ in the quantization function $h(\cdot)$ for notational convenience. In the quantization function  $h(\cdot)$, $\bz_e$ is quantized into $\bz_q$ by assigning a code-vector from the codebook $C$ using a distance measure $d(\cdot, \cdot)$:

\vspace{-0.15in}
\begin{align}
\mathbf{z}_q= \mathbf{c}_k, \quad \text{where} \quad k = \argmin_j d( \bz_e, \mathbf{c}_j) 
\end{align}
\vspace{-0.15in}

Euclidean distance is the standard distance measure for $d(\cdot, \cdot)$~\cite{van2017neural}. We denote the set associated to $\bz_e$, $\bz_q$ and $\bc$ with $\mathcal{P}_{\bz}$, $\mathcal{Q}_{\bz}$ and $\mathcal{C}_{\bz}$, respectively, with $\mathcal{Q}_{\bz} \subseteq \mathcal{C}_{\bz}$. Without loss of generality, we assume these sets are constructed from an underlying distribution.

The quantized embedding is then used to predict the output $\hat \by = G( \bz_q)$, and the loss is computed with the target $\by: \mathcal{L}\left( \hat \by, \by \right)$. 
For images, VQ is generally performed on each spatial location of the tensor, where the channel dimension is used to represent the vector (\eg each spatial location $(i, j)$ of $\bz_e \in \mathbf{R}^{h \times w \times c}$ is quantized \smash{$\bz_e[i,j] \in \mathbf{R}^{c} \xrightarrow[]{ _{h(\cdot)}} \bz_q[i,j] \in \mathbf{R}^{c}$}). 

Akin to standard training, the objective of VQNs is to minimize the empirical risk:

\vspace{-0.15in}
\begin{align}
\min_{F, G, h} \mathbb{E}_{(\bx, \by) \sim \mathcal{D}} \left[ \mathcal{L}_{\mathsf{task}} \left( G( h( F(\bx))), \by \right) \right] 
\end{align}
\vspace{-0.15in}

The equation above is not continuously differentiable. To differentiate through the $\argmin$ operator in $h(\cdot)$, a \textit{straight-through} estimation~\cite{bengio2013estimating} is applied:

\vspace{-0.15in}
\begin{align}
\frac{\partial \cL}{\partial F} =
\frac{\partial \cL}{\partial \hat \by}
\frac{\partial \hat \by}{\partial \bz_q} 
\underbracket[0.140ex]{\cancel{\frac{\partial \bz_q}{\partial \bz_e}}}_{\scriptsize \clap{$\mathsf{straight\;through}$}}
\frac{\partial \bz_e}{\partial F} 
\approx \hat{ \frac{\partial \cL}{\partial F}}
\end{align}
\vspace{-0.15in}

To ensure the straight-through estimation is accurate, the codebook and the encoder representations are pulled together using a \textit{commitment loss}: 

\vspace{-0.15in}
{\small
\begin{align}
\mathcal{L}_{\mathsf{cmt}}(\bz_e, \bz_q) = 
 (1 - \beta) & \cdot d \left(\bz_e, \mathsf{stop\_gradient}\left[\bz_q\right]) \right) \\ 
 + \beta & \cdot d \left(\mathsf{stop\_gradient} \left[ \bz_e \right], \bz_q \right)
\end{align}
}
\vspace{-0.15in}

Here, $\beta \in [0, 1]$ is a scalar that trades off the importance of updating $\bz_e$ and $\bz_q$ (\eg large $\beta$ implies more emphasis on the codebook to adapt towards the encoder). 
Then for some scalar $\alpha$ that weighs the commitment loss, a differentiable pseudo-objective is minimized:

\vspace{-0.15in}
{\small
\begin{align}
\hspace{-0.1in} \min_{F, G, h} \mathbb{E}_{(\bx, \by) \sim \mathcal{D}} \left[ \mathcal{L}_{\mathsf{task}} \left( G( h( F(\bx))), \by \right)  + \alpha \cdot \mathcal{L}_{\mathsf{cmt}}(\bz_e, \bz_q) \right]
\label{eqn:vq-obj}
\end{align}
}
\vspace{-0.15in}

A good rule of thumb is to set $\alpha=10$ and $\beta=0.9$ when using euclidean distance $d_{l_2}(\bz_e, \bz_q):= \frac{1}{2}\lVert \bz_e - \bz_q \rVert_2^2$~\cite{van2017neural}. \\

\xpar{Updating codebook using EMA}
Instead of using the commitment loss, another popular approach is to use exponential moving average~(EMA) to train the codebook: 

\vspace{-0.15in}
\begin{align}
\bz_q^{(t+1)} \leftarrow (1 - \gamma ) \cdot \bz_q^{(t)} + \gamma \cdot \bz_e^{(t)}
\label{eqn:ema}
\end{align}
\vspace{-0.15in}

While EMA is proposed as a training trick in lieu of commitment loss, it is easy to see that it is equivalent to the commitment loss optimized with SGD when $\beta=1$ (see~\app{app:ema}); where the EMA decay constant is the learning rate $\gamma = \eta$. This equivalence has been commonly overlooked with the exception of~\cite{lancucki2020robust}.

\begin{figure*}[!htb]
     \centering
     \includegraphics[width=1.0\linewidth]{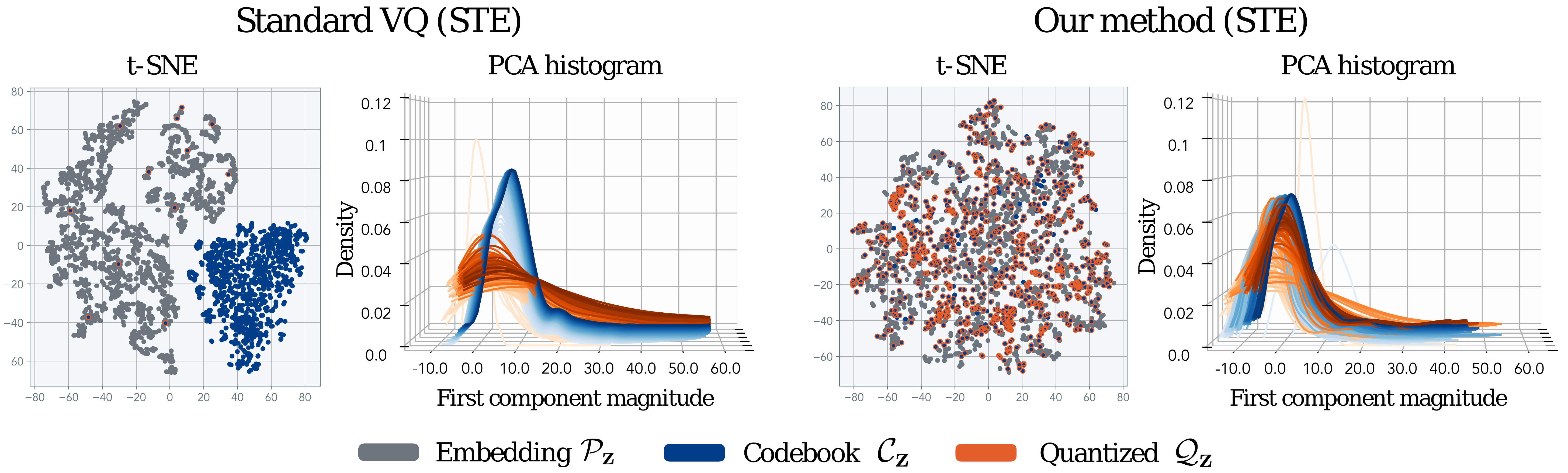}
     \caption{
     \small
     \textbf{Divergence vs Accuracy:} 
     We visualize the divergence between $\mathcal{P}_z$, $\mathcal{Q}_z$ and $\mathcal{C}_{\bz}$ on ResNet18 during training. ResNet18 is trained to solve ImageNet100 classification and the codes are initialized using K-means. We sample $2048$ embedding the vectors from $\mathcal{P}_{\bz}$ and use the full $\mathcal{C}_{\bz}$ and $\mathcal{Q}_{\bz}$. We embed the vectors associated to the 10th training iteration using tSNE. We also compute distribution shifts throughout training by computing the histogram on the PCA projections. Here the lighter color indicates early iteration in training. The standard approach results in a bifurcation of the codebook. On the right, we show the result of our method using affine re-parameterization, which leads to better distribution matching. 
     }
     \label{fig:divergence}
\end{figure*}

\section{On the trainability of VQ networks}
\label{sec:sec3}
It is well known that VQNs perform poorly when the number of actively used codes is small~\cite{kaiser2018fast}. This is referred to as ``index collapse'' and is the bottleneck in training VQNs. 
Thus, there have been abundant efforts to construct algorithms that recover and prevent models from collapsing. We discuss a few popular approaches below:

\xpar{Stochastic sampling} 
\citet{roy2018theory,kaiser2018fast,sonderby2017continuous} proposed to use stochastic sampling and probabilistic relaxation to VQ. Since then, it has become a valid alternative method for training VQNs~\cite{williams2020hierarchical,lee2022autoregressive}. Taking one of the most recent work as an example, \citet{takida2022sq} argues that determinism is the main cause of the codebook collapse and proposes to sample codes from a categorical distribution proportional to the negative distance:
\begin{align}
p(\bz_q = \bc_j | \bz_e) \propto e^{-d(\bz_e - \bc_j) / \tau} 
\end{align}
where $\tau$ is a scalar indicating the temperature. The temperature is often annealed to make the model deterministic at convergence. Stochastic sampling can be a bottleneck as it requires computing and storing the full distance matrix. Note that the original idea of sampling to encourage diversity can be rooted back to~\citet{kohonen1990improved}.

\definecolor{Gray}{gray}{0.9}
\newcolumntype{g}{>{\columncolor{Gray}}c}
\begin{table}
\begin{center}
\scalebox{0.75}{
\begin{tabular}{lcccc} 
    \toprule
    
    {\multirow{2}{*}{Method}} & \multicolumn{2}{c}{\bf K-means} & \multicolumn{2}{c}{$\mathcal{N}_{\mathsf{kaiming}}$} \\  
    \cmidrule(lr){2-3} \cmidrule(lr){4-5}
    & Active & Accuracy & Active & Accuracy \\ 
    \midrule
    None                & 81.4 & 67.0        & 6.8  & 59.3 \\
    Stochastic          & 87.8 & 67.3        & 16.3 & 60.8 \\
    Rep. KM             & 99.2 & 63.7 (67.4) & 99.0 & 62.7 (68.6) \\ 
    \rowcolor{Gray} Replace (LRU) & 99.8 & 69.3        & 99.9 & 68.6\\  
\bottomrule
\end{tabular}
}
\caption{\small \textbf{Collapse prevention:} Comparison of index collapse prevention mechanisms proposed in prior works. We report the  classification performance using $\mathsf{ResNet18}$ on $\mathsf{ImageNet100}$ when initializing the code-vectors with K-means and $\mathcal{N}_{\mathsf{kaiming}}$. All models use $1024$ codes with vector-size of $512$.}
\label{table:comparison}
\end{center}
\vspace{-0.1in}
\end{table}

\xpar{Repeated K-means}
\citet{lancucki2020robust} explicitly ensures all code-vectors are active by running K-means at every epoch. The naive implementation of repeated K-means forces all codes to be re-initialized. Depending on the noise sensitivity of the decoder, it can lead to large spikes in model performance, where both the encoder and the decoder have to readjust to the newly introduced codes. When decaying the learning rate, the model can no longer adapt to the new codes, and the performance stars to degrade.

\xpar{Replacement policy}
\citet{zeghidour2021soundstream,dhariwal2020jukebox} proposed to replace the dead codes with a randomly sampled embedding vector. 
Replacement policies require careful tuning, and we find that using a least-recently-used (LRU) policy with a life-span of $20$ iteration works the best -- if the code does not get used for $20$ training iterations, it gets replaced with a random embedding vector.
When using replacement policies, the active codes are left unchanged, and the overall performance of the model does not degrade. 

As shown in~\tbl{table:comparison}, these aforementioned works result in better performance with an improved number of actively used codes.
However, these methods address the symptoms of the collapse by replacing inactive codes rather than resolving why they became inactive in the first place.
In this work, we investigate the source of the index collapse by analyzing the optimization dynamics of the codes and how it affects the model. We find that the divergence in the model representation causes the index collapse, and this divergence causes erroneous model updates.

\subsection{Commitment loss is an asymmetric loss}

VQ layers often diverge throughout training and fail to recover the codes that are no longer actively used (see~\app{app:early-collapse}). While having good initial conditions, such as an improved initialization scheme~(see~\app{app:init}), can improve index collapse, ensuring good codebook usage is hard to maintain throughout training. To understand why dead codes are hard to recover from, we need to revisit the commitment loss. One can rewrite commitment loss as an average over distance $d(\cdot)$ computed between an aligned set of points in $\mathcal{P}_{\bz}$ and $\mathcal{C}_{\bz}$:

\vspace{-0.1in}
{\small
\begin{align}
\min_{\mathcal{C}_{\bz}} D(\mathcal{P}_{\bz} , \mathcal{C}_{\bz}) = \frac{1}{\lvert \mathcal{P}_{\bz} \rvert}\sum_{\bz_i \in \mathcal{P}_{\bz}} \min_{\bc_j \in \mathcal{C}_{\bz}} d(\bz_i, \bc_j) 
\end{align}
}
\vspace{-0.1in}

Here, when $d(\cdot)$ is a Bregman divergence (\eg $l_2$ used in commitment loss is a Bregman divergence), then the distance $D(\mathcal{P}_{\bz}, \mathcal{C}_{\bz})$ can be seen as an average divergence over an aligned set of $\mathcal{P}_{\bz}$ and $\mathcal{C}_{\bz}$~\cite{banerjee2005clustering}. 

This divergence function is non-symmetric and is computed over $\mathcal{P}_{\bz}$ but is minimized with respect to $\mathcal{C}_{\bz}$. Rewriting the commitment loss as an average divergence makes it easier to see why it is susceptible to model collapse. The divergence results in a many-to-one mapping, with the set of selected codes forming $\mathcal{Q}_{\bz}$. Here the disjoint set $\mathcal{C}_{\bz} \setminus \mathcal{Q}_{\bz}$ does not receive any gradient and is not trained. This is analogous to computing reverse-KL in probability measure~\cite{ghosh@revserkl}, where any sample that falls outside the support of the measure $\mathcal{P}$, does not receive gradients (see~\app{app:cmt} for further discussion). 

Since the subset of codes $\mathcal{Q}$ is used to minimize the commitment loss, and \textit{not} $\mathcal{C}$, $\mathcal{Q}$ learns a ``mode-seeking'' behavior. This implies that once the codes are not selected, they will likely remain unselected in the future. Note that even if we initialize the code-vectors to overlap in distribution with the embedding perfectly, the code-vectors can be dropped during optimization for various reasons, including stochasticity in training and non-stationary model representation $\mathcal{P}_{\bz}$. 

We further demonstrate the impact of the divergence between the codebook and encoder embeddings by visualizing how they diverge in practice. In~\fig{fig:divergence}~(left), we train ResNet18~\cite{he2016deep} on the ImageNet100~\cite{russakovsky2015imagenet,tian2020contrastive} dataset, initialized with K-means, and visualize $\mathcal{P}{\bz}$, $\mathcal{Q}{\bz}$, and $\mathcal{C}_{\bz}$ using dimension reduction methods after several optimization steps. Using t-SNE, we observe that only a few code-vectors are active, with more than $95\%$ of the codes not being selected and trained. We also compute the histogram of the vectors by projecting them into the first PCA component. The histogram shows that the distribution quickly diverges after a few iterations of training.
The experiment highlights the vulnerability of VQN optimization, where a sudden shift in the encoder embedding causes severe misalignment. Once misaligned, it often stays misaligned throughout optimization, with few active codes representing the embedding representation.
A more desirable outcome can be observed in~\fig{fig:divergence}~(right), where the codebook can closely match the embedding distribution. We discuss how to achieve better distribution matching in~\sect{sec:affine}.

\subsection{Gradient estimation gap} 
\label{sec:ste-err}

When the model embedding diverges from the codebook distribution, the quantization function yields sub-optimal code assignments. This sub-optimal assignment results in a sudden increase in the average quantization error~(see~\app{app:early-collapse}). As VQNs rely on straight-through estimation, the accuracy of the gradient updates in the encoder becomes dependent on the precision of the quantization function. 

A good quantization function $h(\cdot)$ is one that can preserve the necessary information of $\bz_e$ given a finite set of vectors. The resulting quantization vector can be represented as $\bz_q = \bz_e + \epsilon$ where $\epsilon$ is the residual error vector resulting from the quantization function. When $\epsilon = \mathbf{0}$, there exists no straight-through estimation error, and the model acts as if there is no quantization function. Of course, with a finite set of codes, a lossless quantization function is non-trivial to achieve when training on a large dataset. To measure the gradient deviation from the lossless quantization function, we define the gradient gap as:

\vspace{-0.15in}
\begingroup\makeatletter\def\f@size{9}\check@mathfonts
\begin{align}
 \Delta_\mathsf{gap}(F | h) &= \left\lVert \frac{\partial \mathcal{L}_{\mathsf{task}}(G(\bz_e))}{\partial F(\bx)} - \frac{\partial \mathcal{L}_{\mathsf{task}}(G(\bz_e + \epsilon))}{\partial F(\bx)}  \right\rVert 
 \label{eqn:grad-err}
\end{align}
\endgroup
\vspace{-0.15in}

The gradient gap measures the difference between the gradient of the non-quantized model and the quantized model. When $\Delta_{\mathsf{gap}}=\mathbf{0}$, the gradient descent using STE is guaranteed to minimize the loss. When the gap is large, no guarantees can be made. This gradient gap can be made small when (1) the quantization error is small and (2) the decoder function $G(\cdot)$ is smooth.
To see this, consider the case when $\bz_e$ and $\bz_q$ are equivalent~$(\epsilon=\mathbf{0})$, then the estimation gap is $\Delta_\mathsf{gap}=\mathbf{0}$. When they are not equivalent and $G(\cdot)$ is $K$-Lipschitz smooth, then the estimation error is proportionately bounded by the quantization error $K \cdot d(\bz_e, \bz_q)$. 

While regularizing for the smoothness of the network is task and model-dependent, the quantization error is a controllable design choice that users can improve upon. 
One approach is to ensure the quantization error is small at initialization using K-means~\cite{lancucki2020robust,zeghidour2021soundstream,dvq2021karpathygithub}(see~\app{app:init}). 
Another approach is to further improve gradient estimates by ensuring the quantization error is small throughout training. This can be done by improving the optimization algorithm itself. In~\sect{sec:alt-opt}, we propose an improved optimization algorithm to reduce the gradient estimation gap.

The gradient gap provides a measure of the goodness of the STE and is a useful tool for mathematical intuition. However, there is a caveat to be aware of when using it in practice. When VQNs go through index-collapse, there is a sharp spike in $\Delta_{\mathsf{gap}}$ but eventually, it becomes very trivial for the model to achieve $\Delta_{\mathsf{gap}} = 0$ as the encoder function is encouraged to predict the few remaining codes via the commitment loss. Where with fewer active codes, the easier it becomes to predict. Hence, one should be cautious when using the gradient gap as a measure to compare models.

\begin{figure*}[!htb]
     \centering
     \includegraphics[width=1.0\linewidth]{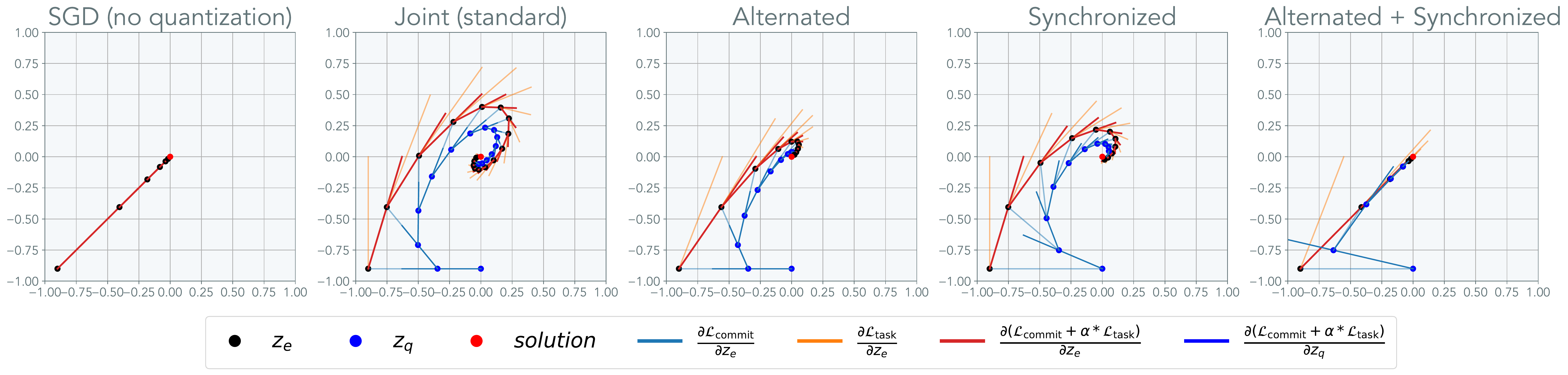}
     \caption{
     \small
     \textbf{Codebook update dynamics on toy experiment:} 
     Optimization dynamics of vector-quantization on a toy setup. The experiment above uses a single code vector with a stationary target for $\bz_e$~(\red{red}). A euclidean loss is computed with respect to the code-vector $\bz_q$~(\blue{blue}), and the resulting gradient is used to update the embedding $\bz_e$ using the straight-through approximation~(black). A commitment loss is applied to the $\bz_q$ and $\bz_e$ using $l_2$ distance. All methods are optimized using SGD with the same fixed learning rate of $0.1$. $\mathsf{No VQ}$ optimizes the embedding without the quantization function. $\mathsf{Joint}$ optimizes the embedding with $\mathcal{L}_{\mathsf{task}}$ and $\mathcal{L}_{\mathsf{commit}}$ together. $\mathsf{Alternated}$ uses first optimizes the codebook assignment and then optimizes the model with the task loss, with single iteration for each step. $\mathsf{Lookahead}$ objective predicts the trajectory of $\bz_e$ and updates $\bz_q$ towards it. The trajectory creates a large spiral for standard VQ training due to the straight-through approximation error. The alternated optimization minimizes this approximation error, reducing the extent of the spiral. Note that the approximation error is also caused by the code-vector representation, which is a historical moving average of the embedding with a delay. The lookahead optimizer reuses the gradient from $\mathcal{L}_{\mathsf{task}}$ to better synchronize the code-vector representation and accelerate convergence.} 
     \label{fig:traj}
\end{figure*}

\section{Improved techniques for VQNs}
The previous section emphasized contributing factors of index collapse and how minimizing the divergence at initialization leads to improved performance and codebook usage. This section explores methods to reduce codebook divergence and improve optimization.

\subsection{Minimizing internal codebook covariate shift with shared affine parameterization}
\label{sec:affine}

In~\sect{sec:sec3}, we observed that methods such as resampling resulted in improved performance with a higher number of active codes. We hypothesize that replacement methods work well because the model representation $\mathcal{C}_{\bz}$ eventually ends up resembling the representation of $\mathcal{P}_{\bz}$ by consistently resampling code-vectors, albeit a slow process that requires resampling at almost every iteration. In light of this observation, we propose a more efficient method to match the distribution of $\mathcal{P}_{\bz}$. But first, we describe why misalignment occurs in the first place.

The misalignment between the internal representation of the network's layers is referred to as an internal covariate shift. VQ layers are also prone to this internal covariate shift, where the consistently changing internal representation $\mathcal{P}_{\bz}$ creates a misalignment with the codebook distribution $\mathcal{C}_{\bz}$. We refer to this misalignment as \textit{internal codebook covariate shift}. Generally, internal covariate shifts can be minimized by directly matching the moments of the distributions, or in the case of \cite{ioffe2015batch}, the distributions are whitened to a univariate Gaussian. 

Using vector-quantization adds a layer of complexity that compounds with the existing internal covariate shift. While the linear layers receive dense gradients from the objective, the codebook receives sparse gradients. This implies that when the internal representation $\mathcal{P}_{\bz}$ is updated, not only does it require much longer for $\mathcal{C}_{\bz}$ to catch up but also if the update in $\mathcal{P}_{\bz}$ is too large~(\eg large learning rates), the assignment can be severely misaligned~(see~\fig{fig:divergence} and~\app{app:affine-toy}).

To reduce gradient sparsity and encourage the codebook to update faster towards the embedding, we propose an affine reparameterization of the code-vector with a shared global mean and standard deviation.

\vspace{-0.15in}
\begin{align}
\bc^{(i)} = \bc_{\mathsf{mean}} + \bc_{\mathsf{std}} * \bc^{(i)}_{\mathsf{signal}}
\end{align}
\vspace{-0.15in}

Here $\bc^{(i)}_{\mathsf{signal}}$ is the original code-vector, and $\bc_{\mathsf{mean}}, \bc_{\mathsf{std}}$ are the shared affine parameters with the same $\text{dim}(\bc^{(i)}_{\mathsf{signal}})$. 

The affine parameters can either be learned through gradient descent or computed via the exponential moving average over $\bz_e$ and $\bz_q$ statistics~(see~\app{app:aff-ema}). Note, that under the Gaussian assumption on $\mathcal{P}_{\bz}$ and $\mathcal{C}_{\bz}$, matching the moments is equivalent to minimizing the KL divergence~\cite{kurz2016kullback}. The reparameterization allows gradients to flow through the unselected code-vectors through the affine parameters. Although we make no specific Gaussian assumption on the parameterization, it is easy to extend our method to better capture complex distributions by assigning distinct affine parameters to each codebook subset.

\subsection{Alternated optimization}
\label{sec:alt-opt}

In~\sect{sec:ste-err}, we showed that the error in the gradient update is proportional to the quantization error. Hence, we are interested in ensuring that divergence stays well-behaved during optimization. Here it is important to note that when there is an index collapse, the model trivially achieves zero gradient error as the commitment loss is easily minimized. This is not the setting we are interested in, and we assume that we are operating on a well-behaved regime. 

For any arbitrary task $\mathcal{L}_{\mathsf{task}}$, the underlying objective of a VQN is to minimize the empirical loss while learning a good codebook representation.

\vspace{-0.15in}
\begin{align}
\min_{F, G, h} \mathbb{E}_{(\bx, \by) \sim \mathcal{D}_{\mathsf{train}}} \left[ \mathcal{L}_{\mathsf{task}} \left( G( h( F(\bx))), \by \right) \right] 
\end{align}
\vspace{-0.15in}

The objective function above is not continuously differentiable; therefore, \eqn{eqn:vq-obj} is used as a surrogate objective. However, the gradient computed from the surrogate objective is a biased estimate of the true gradient and can result in undesirable optimization dynamics. This is illustrated in~\fig{fig:traj} with a toy setup. The dynamical error induced from the optimization can be traced to the straight-through estimation, in which the surrogate gradient deviates from the true gradient proportional to the quantization error. Hence, updating the network when the quantization error is big can lead to an erroneous model update.
To reduce the quantization error, we propose an alternating optimization algorithm:

\vspace{-0.15in}
\begin{align}
&\min_{h} \hspace{0.in} \underset{(\bx, \by) \sim p_{\mathsf{data}}}{\mathbb{E}} \hspace{-0.1in} \left[ \mathcal{L}_{\mathsf{commit}} \left( \textcolor{teal}{h} ( F(\bx)), F(\bx) \right) \right] \label{eqn:alt1}\\
&\min_{F, G} \hspace{0.in} \underset{(\bx, \by) \sim p_{\mathsf{data}}}{\mathbb{E}} \hspace{-0.1in} \left[ \mathcal{L}_{\mathsf{task}} \left( \textcolor{teal}{G}( h( \textcolor{teal}{F}(\bx))), \by \right) \right]
\label{eqn:alt2}
\end{align}
\vspace{-0.15in}

The algorithm above resembles that of online K-means with a non-linear encoder and decoder. Where \eqn{eqn:alt1} optimizes the K-mean clusters, and \eqn{eqn:alt2} optimizes the model given the new cluster assignment.
We know that when $\mathcal{L}_{\mathsf{commit}} \rightarrow 0$, $h(\cdot)$ acts as an identity function under stationary $F$ and $G$. Then, under fixed $h$, both $F$ and $G$ can be optimized with close to zero estimation error. This can be repeated until the small quantization error assumption is broken. 

Of course, optimizing the inner term till convergence is computationally expensive. Fortunately, we find that it is not necessary to wait till convergence (see~\app{app:alt-opt}). In practice, alternating the inner term even for a single iteration yields good performance. In the toy setting of~\fig{fig:traj}, we visualize how the alternating optimization performs when using a single update for each term. 

\subsection{One-step behind or in synchronous step?}
The codebook updated with the commitment loss is a historical average of the model representation. 
Writing out the gradient update for commitment loss for $\beta=1$:

\vspace{-0.15in}
\begin{align} %
\bz_q^{(t+1)} \leftarrow (1 - \eta ) \cdot \bz_q^{(t)} + \eta \cdot \bz_e^{(t)} 
\end{align}
\vspace{-0.15in}

The historical average does not account for the current representation but only up to the previous representation. Therefore, computing the gradient with respect to the historical average implies that the model receives a ``delayed'' gradient.
To reduce the delay in the $\bz_q$ representation, we desire the code-vector to include a running average of the most recent representation:

\vspace{-0.15in}
\begin{align}
\bz_q^{(t+1)} \leftarrow (1 - \eta ) \cdot \bz_q^{(t)} + \eta \cdot \bz_e^{\textcolor{teal}{(t+1)}}
\end{align}
\vspace{-0.15in}

The above equation is tractably computable as the gradient of $\bz_q$ is used to update $\bz_e$. Hence, an explicit equation for the \textit{synchronized update rule} is:

\vspace{-0.15in}
\begin{align}
\bz_q^{(t+1)} \leftarrow \underbracket{(1 - \eta ) \cdot \bz_q^{(t)} + \eta \cdot \bz_e^{(t)}}_{\text{original commitment loss}} + \textcolor{teal}{\eta^2 \cdot \frac{\partial \mathcal{L}_{\mathsf{task}}}{\partial \bz_q}}
\end{align}
\vspace{-0.15in}

Using the equation above, the code-vectors take a step in the direction of the encoder representation using the gradient of the task loss. In python, this requires a minor change to the existing implementation of straight-through estimation:

\vspace{-0.15in}
{\small
\begin{equation}
\mathtt{z\_q= z + (z\_q - z).detach() + \textcolor{teal}{\nu * (z\_q - (z\_q).detach())}}
\nonumber
\end{equation}
}
\vspace{-0.15in}

Where $\nu$ is a scalar to decide whether we want a pessimistic or an optimistic update. We find that the effectiveness of $\nu$ depends on the model architecture.

\section{Results}

\subsection{Classification} 
We apply our methods to ImageNet100~\cite{tian2020contrastive} classification. ImageNet100 is a subset of the ImageNet-1K dataset~\cite{russakovsky2015imagenet}, consisting of 100 classes and approximately 100,000 images. The training details are provided in the~\app{app:train-details}.

Our results, presented in~\tbl{table:classification}, demonstrate the effectiveness of our proposed methods. All models were initialized using the K-means clustering algorithm. We also report the perplexity of the model on the test dataset, which is defined as $2^{H(p)}$, where $H(p)$ is the entropy over the codebook likelihood. A higher perplexity implies a uniform assignment of codes. While having very low perplexity is associated with index collapse, having more does not necessarily imply better performance -- having a high perplexity on a task that has more codes than necessary indicates redundancy. Our results using affine re-parameterization largely improves index-collapse, and the use of synchronized and alternating training methods further improves the overall performance of the model. We further compare our results against the use of the least-recently-used (LRU) replacement policy and observed our method to outperform or perform comparably. A combination of all these methods results in the largest improvement in performance. We observed using $l_2$ normalization to hurt performance on classification. We suspect that removing the magnitude component of the embedding hurts models that use magnitude-sensitive objectives (\eg soft-max cross-entropy loss).

\begin{table}[t!]
\begin{center}
\scalebox{0.7}{
\begin{tabular}{l|ccc|c|cc} 
    \toprule
    & Affine & Sync. & Alt. & Replace & Accuracy $\uparrow$ & Perplexity $\uparrow$ \\
    \midrule
    {\multirow{6}{*}{\rotatebox[origin=c]{90}{$\mathsf{AlexNet}$}}}
    & - &  - & - &   - & 47.2 & 11.2\\  
    & \checkmark &  - & - &   - & 55.7 (+8.5) & 648.1 \\  
    & - &  \checkmark & - &   - & 49.2 (+2.0) & 10.7\\  
    & - &  - & \checkmark &   - & 52.8 (+5.6) & 65.7\\  
    & - &  - & - &   \checkmark & 54.4 (+7.2) & 657.6\\  
    & \checkmark & \checkmark & \checkmark & \checkmark & \textbf{57.9 (+10.7)} & \textbf{753.8}\\  
    \midrule
    {\multirow{6}{*}{\rotatebox[origin=c]{90}{$\mathsf{ResNet18}$}}}
    & - &  - & - &   - & 64.1 & 32.4 \\  
    & \checkmark &  - & - &   - & 70.4 (+6.3) & 300.0\\  
    & - &  \checkmark & - &   - & 66.0 (+1.9) & 60.1\\  
    & - &  - & \checkmark &   - & 67.5 (+3.4) & 54.2\\  
    & - &  - & - &   \checkmark & 68.1 (+4.0) & \textbf{334.5} \\  
    & \checkmark & \checkmark & \checkmark & \checkmark & \textbf{71.0 (+6.9)} & 306.6 \\  
    \midrule
    {\multirow{6}{*}{\rotatebox[origin=c]{90}{$\mathsf{ViT(T)}$}}}
    & - &  - & - &   - & 48.6 & 112.9\\  
    & \checkmark &  - & - &   - & 52.8 (+4.2) & 299.0\\  
    & - &  \checkmark & - &   - & 51.3 (+2.7) & 88.8\\  
    & - &  - & \checkmark &   - & 53.1 (+4.5) & 95.6\\  
    & - &  - & - &   \checkmark & 54.4 (+5.8) & \textbf{598.8}\\  
    & \checkmark & \checkmark & \checkmark & \checkmark & \textbf{56.7 (+8.1)} & 592.2\\
\bottomrule
\end{tabular}
 }
\caption{\small \textbf{Classification:} The effect of how our methods affect the final performance on classification. All methods are initialized with K-means.}
\label{table:classification}
\end{center}
\end{table}

\subsection{Generative modeling}
We further apply our method to CelebA~\cite{liu2015faceattributes} and CIFAR10~\cite{krizhevsky2009learning} generative modeling tasks. 
We adopt the training framework from previous works, and the details can be found in the appendix~(\app{app:train-details}). We compare the performance of our method against existing techniques such as VQVAE~\cite{van2017neural}, SQVAE~\cite{takida2022sq}, and Gumbel-VQVAE~\cite{dvq2021karpathygithub,esser2020taming} using MSE as well as LPIPS perceptual loss~\cite{zhang2018perceptual}. SQVAE requires $4\times$ more memory footprint than all other methods as it requires storing the full distance matrix along with the computation graph. Both baselines using $l_2$ normalization and least-recently-used (LRU) replacement policy largely improve training stability and reconstruction performance of generative models. When these methods are applied jointly with ours, we observe the best improvement.

\begin{figure}[t!]
     \centering
     \includegraphics[width=0.495\linewidth]{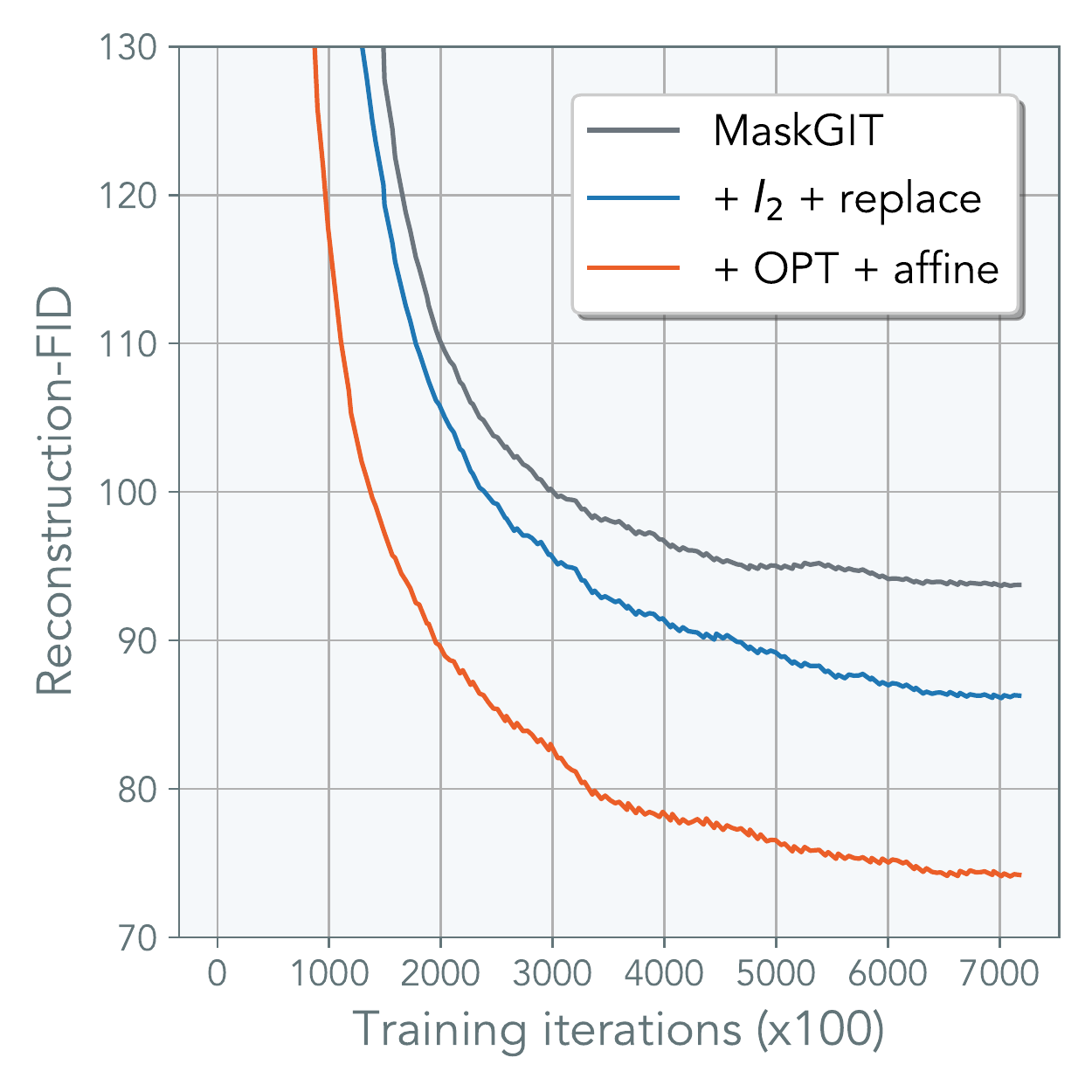}
     \hfill
     \includegraphics[width=0.495\linewidth]{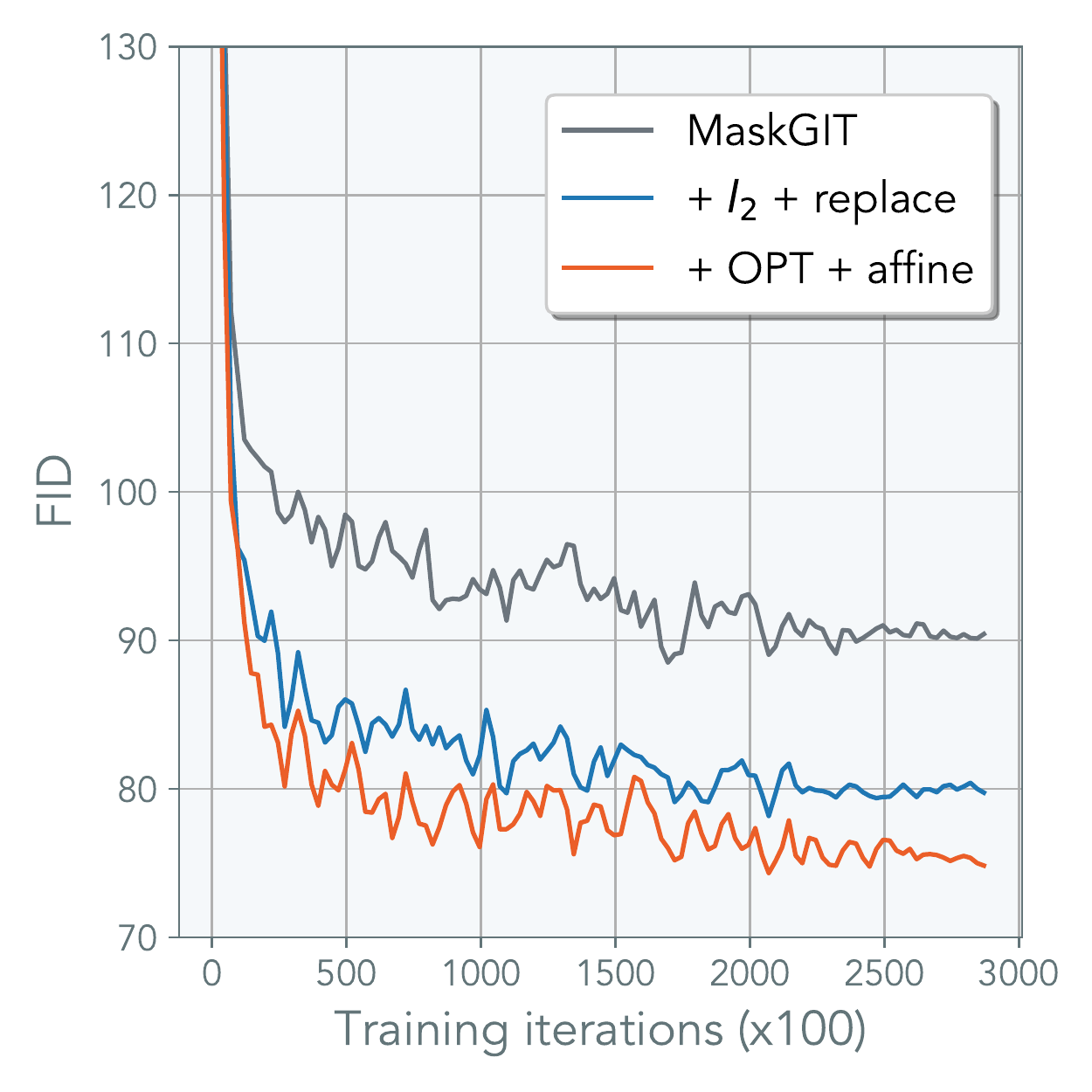}
     \caption{\small\textbf{MaskGIT FID training curves:} MaskGIT~\cite{chang2022maskgit} trained on CelebA with only reconstruction loss. We report rFID~(left) and FID~(right) training curves. We use a slimmed-down version of MaskGIT: VQGAN using $32$ channels instead of $128$ and transformer using $8$ blocks instead of $24$.}
     \label{fig:maskgit-fid}
\end{figure}

In~\fig{fig:maskgit-fid}, we run generative modeling using MaskGIT~\cite{chang2022maskgit}. We plot the rFID (reconstruction-FID)~\cite{takida2022sq} and the FID~\cite{heusel2017gans} during training. rFID measures the FID on the reconstructed images from the auto-encoder over the test set. We do not use perceptual or discriminator loss to train the network, and for rFID/FID, we use $5000$ generated samples.

\subsection{Warmup and normalization can be helpful}
\label{sec:warmup}

One way to mitigate the divergence between the codebook and the embedding distribution is by constraining the representation to be within a bounded measure space. This limits the range of movement of the embedding distribution, facilitating alignment between the codebook and the embedding distribution throughout training. Common techniques for this include $l_2$ normalization~\cite{yu2021vector}, batch-normalization~\cite{lancucki2020robust}, and assuming a restricted distribution~\cite{takida2022sq} in probabilistic VQNs. However, these techniques improve stability at the cost of reduced model expressivity. Alternatively, one can ensure the updates of $\mathcal{P}_{\bz}$ to be small in order for the codebook to catch up. To do so without hindering convergence speed, we find a learning rate scheduler with warmup to work very well. In~\app{app:warmup}, we show how using cosine learning rate decay with linear warmup~\cite{loshchilov2016sgdr,goyal2017accurate} improves both the model performance and model perplexity.

\subsection{Ablation on alternating optimization}
In~\app{app:alt-opt}, we measure how varying the number of inner and outer loop iterations affects classification performance. We find that by increasing the number of inner loop iterations by $8$, we observed an $11.09\%$ improvement over the baseline and $5.51$\% over the version that uses a single iteration of the inner step. On the other hand, we do not find increasing the outer loop to help. When combining all our methods, we observed that setting the inner loop iteration to 1-2 suffice. In this experiment, we evenly divide the training mini-batch into sub-mini-batches for each iteration of the expectation and maximization steps. This ensures that the number of images the model observes is equal across all experiments. Furthermore, when choosing to only optimize $h$ for a single iteration, it is possible to update both the inner and outer term in a single forward pass, as the commitment loss $\mathcal{L}_{\mathsf{commit}}$ does not depend on the task loss $\mathcal{L}_{\mathsf{task}}$. As a result, the computational overhead for a single fused pass is $1.05\times$, and $2\times$ with four inner loop iterations (For smaller models like AlexNet, the overhead is less, roughly $1.5\times$).

\begin{table}[t!]
\begin{center}
\scalebox{0.6}{
\begin{tabular}{l|l|ccc} 
    \toprule
     & Method & MSE ($10^{-3}$) $\downarrow$ & Perplexity $\uparrow$ & LPIPS ($10^{-1}$) $\downarrow$ \\
    \midrule
    {\multirow{11}{*}{\rotatebox[origin=c]{90}{$\mathsf{CIFAR10}$}}}
    & VQVAE                                                  & 5.65 & 14.0  & 5.43  \\  
    & VQVAE + $l_2$                                          & 3.21 & 57.0  & 3.64  \\  
    & VQVAE + replace                                        & 4.07 & 109.8 & 4.30 \\  
    & VQVAE + $l_2$ + replace                                & 3.24 & 115.6 & 3.56 \\  
    & SQVAE                                                  & 3.36 & \textbf{769.3}  & 3.99 \\  
    & Gumbel-VQVAE                                           & 6.16 & 20.3  & 5.73 \\  
    \cmidrule(lr){2-5}
    & VQVAE + Affine                                         & 5.15 & 69.5  & 5.18 \\  
    & VQVAE + OPT                                            & 4.73 & 15.5  & 4.82 \\  
    & VQVAE + Affine + OPT                                   & 4.00 & 79.3  & 4.36 \\  
    & VQVAE + Affine + OPT + replace         & 1.81 & 290.9 & 2.56 \\  
    & VQVAE + Affine + OPT + replace + $l_2$ & \textbf{1.74} & 608.6 & \textbf{2.27} \\  
    
    \midrule
    {\multirow{11}{*}{\rotatebox[origin=c]{90}{$\mathsf{CELEBA}$}}}
    & VQVAE                                                  & 10.02 & 16.2 & 2.71 \\  
    & VQVAE + $l_2$                                          & 6.49 & 188.7 & 1.82 \\  
    & VQVAE + replace                                        & 4.77 & 676.4 & 1.55 \\  
    & VQVAE + $l_2$ + replace                                & 4.93 & 861.7 & 1.47 \\  
    & SQVAE                                                  & 9.17 & 769.1  & 2.68 \\  
    & Gumbel-VQVAE                                           & 7.34 & 96.7  & 2.30 \\  
    \cmidrule(lr){2-5}
    & VQVAE + Affine                                         & 7.47 & 112.6  & 2.22 \\  
    & VQVAE + OPT                                            & 7.78 & 30.5  & 2.25 \\  
    & VQVAE + Affine + OPT                                   & 6.60 & 186.6 & 1.82 \\  
    & VQVAE + Affine + OPT + replace         & \textbf{3.84} & 650.4 & \textbf{1.35} \\  
    & VQVAE + Affine + OPT + replace + $l_2$ & 4.42 & \textbf{872.6} & 1.36 \\  
\bottomrule
\end{tabular}
}
\caption{\small \textbf{Generative modeling reconstruction:} Comparison between various methods on image reconstruction task. All methods use the same base architecture used in~\cite{takida2022sq}. The metrics are computed on the test set and hyper-parameters are tuned for each model.}
\label{table:generative}
\end{center}
\vspace{-0.2in}
\end{table}

\subsection{Further reducing sparsity in VQNs}
Irrespective of the initial alignment between the codebook $\mathcal{C}_z$ and the embedding distribution $\mathcal{P}_z$, a certain degree of divergence between these distributions is inevitable. This is particularly true for loss functions that are either unbounded (e.g., hinge loss) or have non-saturating gradients (e.g., logistic/exponential losses), where the weights grow inversely proportional to the loss. This effect is more pronounced in networks that lack normalization. Despite the use of affine reparameterization of the code-vectors, achieving $100$\% utilization is non-trivial. To further reduce the sparsity in the codebook update, one can directly improve the architectural design choices that contribute to sparsity. Specifically, factors such as image size, batch size, and the number of pooling layers have a significant effect on VQN performance, as the number of code-vector selections directly depends on these variables. In~\app{app:sensitivity}, we demonstrate that there is a significant degradation in performance when reducing the image size from $256\times256$ to $128\times128$, resulting in an over $20$\% reduction in performance. This indicates the importance of training design choices is VQN.

\section{Conclusion}

Discretization has played a significant role in many fields, such as analog-to-digital communication and modern computing. Once representations are made discrete, various techniques from information theory can be applied to manipulate them for benefits such as compression, error correction, and robustness. Discrete representations can also be broken down into independent parts, allowing for the development of composable symbolic representations. In this work, we proposed a set of techniques that addresses several known challenges in optimizing vector-quantized models. Through our proposed methods, we were able to demonstrate improved model performance. While symbolic representation learning is still in its early stages, our optimization techniques provide insight for designing better models in the future.

\section{Acknowledgement}
We want to thank our lab members for their helpful feedback. Minyoung Huh was funded by ONR MURI grant N00014-22-1-2740.
Brian Cheung is supported by the Center for Brains, Minds and Machines (CBMM), funded by NSF STC award CCF-1231216.
Minyoung would like to further thank Wei-Chiu Ma, Lucy Chai, and Eunice Lee.

\bibliographystyle{icml2022}
\bibliography{mlbib}

\begin{thebibliography}{42}
\providecommand{\natexlab}[1]{#1}
\providecommand{\url}[1]{\texttt{#1}}
\expandafter\ifx\csname urlstyle\endcsname\relax
  \providecommand{\doi}[1]{doi: #1}\else
  \providecommand{\doi}{doi: \begingroup \urlstyle{rm}\Url}\fi

\bibitem[Asano et~al.(2020)Asano, Rupprecht, and Vedaldi]{asano2019self}
Asano, Y.~M., Rupprecht, C., and Vedaldi, A.
\newblock Self-labelling via simultaneous clustering and representation
  learning.
\newblock In \emph{International Conference on Learning Representations}, 2020.

\bibitem[Banerjee et~al.(2005)Banerjee, Merugu, Dhillon, Ghosh, and
  Lafferty]{banerjee2005clustering}
Banerjee, A., Merugu, S., Dhillon, I.~S., Ghosh, J., and Lafferty, J.
\newblock Clustering with bregman divergences.
\newblock \emph{Journal of machine learning research}, 6\penalty0 (10), 2005.

\bibitem[Bengio et~al.(2013)Bengio, L{\'e}onard, and
  Courville]{bengio2013estimating}
Bengio, Y., L{\'e}onard, N., and Courville, A.
\newblock Estimating or propagating gradients through stochastic neurons for
  conditional computation.
\newblock \emph{arXiv preprint arXiv:1308.3432}, 2013.

\bibitem[Caron et~al.(2018)Caron, Bojanowski, Joulin, and Douze]{caron2018deep}
Caron, M., Bojanowski, P., Joulin, A., and Douze, M.
\newblock Deep clustering for unsupervised learning of visual features.
\newblock In \emph{Proceedings of the European conference on computer vision
  (ECCV)}, pp.\  132--149, 2018.

\bibitem[Caron et~al.(2020)Caron, Misra, Mairal, Goyal, Bojanowski, and
  Joulin]{caron2020unsupervised}
Caron, M., Misra, I., Mairal, J., Goyal, P., Bojanowski, P., and Joulin, A.
\newblock Unsupervised learning of visual features by contrasting cluster
  assignments.
\newblock \emph{Advances in Neural Information Processing Systems},
  33:\penalty0 9912--9924, 2020.

\bibitem[Chang et~al.(2022)Chang, Zhang, Jiang, Liu, and
  Freeman]{chang2022maskgit}
Chang, H., Zhang, H., Jiang, L., Liu, C., and Freeman, W.~T.
\newblock Maskgit: Masked generative image transformer.
\newblock In \emph{Proceedings of the IEEE/CVF Conference on Computer Vision
  and Pattern Recognition}, pp.\  11315--11325, 2022.

\bibitem[Chang et~al.(2023)Chang, Zhang, Barber, Maschinot, Lezama, Jiang,
  Yang, Murphy, Freeman, Rubinstein, et~al.]{chang2023muse}
Chang, H., Zhang, H., Barber, J., Maschinot, A., Lezama, J., Jiang, L., Yang,
  M.-H., Murphy, K., Freeman, W.~T., Rubinstein, M., et~al.
\newblock Muse: Text-to-image generation via masked generative transformers.
\newblock \emph{arXiv preprint arXiv:2301.00704}, 2023.

\bibitem[Chen et~al.(2022)Chen, Hsieh, and Gong]{chen2021vision}
Chen, X., Hsieh, C.-J., and Gong, B.
\newblock When vision transformers outperform resnets without pre-training or
  strong data augmentations.
\newblock In \emph{International Conference on Learning Representations}, 2022.

\bibitem[Chung et~al.(2020)Chung, Tang, and Glass]{chung2020vector}
Chung, Y.-A., Tang, H., and Glass, J.
\newblock Vector-quantized autoregressive predictive coding.
\newblock In \emph{Interspeech}, 2020.

\bibitem[Dhariwal et~al.(2020)Dhariwal, Jun, Payne, Kim, Radford, and
  Sutskever]{dhariwal2020jukebox}
Dhariwal, P., Jun, H., Payne, C., Kim, J.~W., Radford, A., and Sutskever, I.
\newblock Jukebox: A generative model for music.
\newblock \emph{arXiv preprint arXiv:2005.00341}, 2020.

\bibitem[Esser et~al.(2020)Esser, Rombach, and Ommer]{esser2020taming}
Esser, P., Rombach, R., and Ommer, B.
\newblock Taming transformers for high-resolution image synthesis.
\newblock In \emph{Proceedings of the IEEE Conference on Computer Vision and
  Pattern Recognition}, 2020.

\bibitem[Ghosh(2018)]{ghosh@revserkl}
Ghosh, D.
\newblock Kl divergence for machine learning.
\newblock \url{https://dibyaghosh.com/blog/probability/kldivergence.html},
  2018.

\bibitem[Goyal et~al.(2017)Goyal, Doll{\'a}r, Girshick, Noordhuis, Wesolowski,
  Kyrola, Tulloch, Jia, and He]{goyal2017accurate}
Goyal, P., Doll{\'a}r, P., Girshick, R., Noordhuis, P., Wesolowski, L., Kyrola,
  A., Tulloch, A., Jia, Y., and He, K.
\newblock Accurate, large minibatch sgd: Training imagenet in 1 hour.
\newblock \emph{arXiv preprint arXiv:1706.02677}, 2017.

\bibitem[Gray(1984)]{gray1984vector}
Gray, R.
\newblock Vector quantization.
\newblock \emph{IEEE Assp Magazine}, 1\penalty0 (2):\penalty0 4--29, 1984.

\bibitem[He et~al.(2016)He, Zhang, Ren, and Sun]{he2016deep}
He, K., Zhang, X., Ren, S., and Sun, J.
\newblock Deep residual learning for image recognition.
\newblock In \emph{Proceedings of the IEEE conference on computer vision and
  pattern recognition}, pp.\  770--778, 2016.

\bibitem[He et~al.(2022)He, Chen, Xie, Li, Doll{\'a}r, and
  Girshick]{he2022masked}
He, K., Chen, X., Xie, S., Li, Y., Doll{\'a}r, P., and Girshick, R.
\newblock Masked autoencoders are scalable vision learners.
\newblock In \emph{Proceedings of the IEEE/CVF Conference on Computer Vision
  and Pattern Recognition}, pp.\  16000--16009, 2022.

\bibitem[Heusel et~al.(2017)Heusel, Ramsauer, Unterthiner, Nessler, and
  Hochreiter]{heusel2017gans}
Heusel, M., Ramsauer, H., Unterthiner, T., Nessler, B., and Hochreiter, S.
\newblock Gans trained by a two time-scale update rule converge to a local nash
  equilibrium.
\newblock \emph{Advances in neural information processing systems}, 30, 2017.

\bibitem[Ioffe \& Szegedy(2015)Ioffe and Szegedy]{ioffe2015batch}
Ioffe, S. and Szegedy, C.
\newblock Batch normalization: Accelerating deep network training by reducing
  internal covariate shift.
\newblock In \emph{International conference on machine learning}, pp.\
  448--456. PMLR, 2015.

\bibitem[Jang et~al.(2017)Jang, Gu, and Poole]{jang2016categorical}
Jang, E., Gu, S., and Poole, B.
\newblock Categorical reparameterization with gumbel-softmax.
\newblock In \emph{International Conference on Learning Representations}, 2017.

\bibitem[Kaiser et~al.(2018)Kaiser, Bengio, Roy, Vaswani, Parmar, Uszkoreit,
  and Shazeer]{kaiser2018fast}
Kaiser, L., Bengio, S., Roy, A., Vaswani, A., Parmar, N., Uszkoreit, J., and
  Shazeer, N.
\newblock Fast decoding in sequence models using discrete latent variables.
\newblock In \emph{International Conference on Machine Learning}, pp.\
  2390--2399. PMLR, 2018.

\bibitem[Karpathy(2021)]{dvq2021karpathygithub}
Karpathy, A.
\newblock deep-vector-quantization.
\newblock \url{https://github.com/karpathy/deep-vector-quantization}, 2021.

\bibitem[Kohonen(1990)]{kohonen1990improved}
Kohonen, T.
\newblock Improved versions of learning vector quantization.
\newblock In \emph{1990 ijcnn international joint conference on Neural
  networks}, pp.\  545--550. IEEE, 1990.

\bibitem[Krizhevsky et~al.(2009)Krizhevsky, Hinton,
  et~al.]{krizhevsky2009learning}
Krizhevsky, A., Hinton, G., et~al.
\newblock Learning multiple layers of features from tiny images.
\newblock 2009.

\bibitem[Kurz et~al.(2016)Kurz, Pfaff, and Hanebeck]{kurz2016kullback}
Kurz, G., Pfaff, F., and Hanebeck, U.~D.
\newblock Kullback-leibler divergence and moment matching for hyperspherical
  probability distributions.
\newblock In \emph{2016 19th International Conference on Information Fusion
  (FUSION)}, pp.\  2087--2094. IEEE, 2016.

\bibitem[{\L}a{\'n}cucki et~al.(2020){\L}a{\'n}cucki, Chorowski, Sanchez,
  Marxer, Chen, Dolfing, Khurana, Alum{\"a}e, and Laurent]{lancucki2020robust}
{\L}a{\'n}cucki, A., Chorowski, J., Sanchez, G., Marxer, R., Chen, N., Dolfing,
  H.~J., Khurana, S., Alum{\"a}e, T., and Laurent, A.
\newblock Robust training of vector quantized bottleneck models.
\newblock In \emph{2020 International Joint Conference on Neural Networks
  (IJCNN)}, pp.\  1--7. IEEE, 2020.

\bibitem[Lee et~al.(2022)Lee, Kim, Kim, Cho, and Han]{lee2022autoregressive}
Lee, D., Kim, C., Kim, S., Cho, M., and Han, W.-S.
\newblock Autoregressive image generation using residual quantization.
\newblock In \emph{Proceedings of the IEEE conference on computer vision and
  pattern recognition}, 2022.

\bibitem[Liu et~al.(2015)Liu, Luo, Wang, and Tang]{liu2015faceattributes}
Liu, Z., Luo, P., Wang, X., and Tang, X.
\newblock Deep learning face attributes in the wild.
\newblock In \emph{Proceedings of International Conference on Computer Vision
  (ICCV)}, December 2015.

\bibitem[Loshchilov \& Hutter(2017)Loshchilov and Hutter]{loshchilov2016sgdr}
Loshchilov, I. and Hutter, F.
\newblock Sgdr: Stochastic gradient descent with warm restarts.
\newblock In \emph{International Conference on Learning Representations}, 2017.

\bibitem[Ozair et~al.(2021)Ozair, Li, Razavi, Antonoglou, Van Den~Oord, and
  Vinyals]{ozair2021vector}
Ozair, S., Li, Y., Razavi, A., Antonoglou, I., Van Den~Oord, A., and Vinyals,
  O.
\newblock Vector quantized models for planning.
\newblock In \emph{International Conference on Machine Learning}, pp.\
  8302--8313. PMLR, 2021.

\bibitem[Paszke et~al.(2019)Paszke, Gross, Massa, Lerer, Bradbury, Chanan,
  Killeen, Lin, Gimelshein, Antiga, Desmaison, Kopf, Yang, DeVito, Raison,
  Tejani, Chilamkurthy, Steiner, Fang, Bai, and Chintala]{NEURIPS2019_9015}
Paszke, A., Gross, S., Massa, F., Lerer, A., Bradbury, J., Chanan, G., Killeen,
  T., Lin, Z., Gimelshein, N., Antiga, L., Desmaison, A., Kopf, A., Yang, E.,
  DeVito, Z., Raison, M., Tejani, A., Chilamkurthy, S., Steiner, B., Fang, L.,
  Bai, J., and Chintala, S.
\newblock Pytorch: An imperative style, high-performance deep learning library.
\newblock In Wallach, H., Larochelle, H., Beygelzimer, A., d~Alch\'{e}-Buc, F.,
  Fox, E., and Garnett, R. (eds.), \emph{Advances in Neural Information
  Processing Systems 32}, pp.\  8024--8035. Curran Associates, Inc., 2019.

\bibitem[Ramesh et~al.(2021)Ramesh, Pavlov, Goh, Gray, Voss, Radford, Chen, and
  Sutskever]{ramesh2021zero}
Ramesh, A., Pavlov, M., Goh, G., Gray, S., Voss, C., Radford, A., Chen, M., and
  Sutskever, I.
\newblock Zero-shot text-to-image generation.
\newblock In \emph{International Conference on Machine Learning}, pp.\
  8821--8831. PMLR, 2021.

\bibitem[Roy et~al.(2018)Roy, Vaswani, Neelakantan, and Parmar]{roy2018theory}
Roy, A., Vaswani, A., Neelakantan, A., and Parmar, N.
\newblock Theory and experiments on vector quantized autoencoders.
\newblock \emph{arXiv preprint arXiv:1805.11063}, 2018.

\bibitem[Russakovsky et~al.(2015)Russakovsky, Deng, Su, Krause, Satheesh, Ma,
  Huang, Karpathy, Khosla, Bernstein, et~al.]{russakovsky2015imagenet}
Russakovsky, O., Deng, J., Su, H., Krause, J., Satheesh, S., Ma, S., Huang, Z.,
  Karpathy, A., Khosla, A., Bernstein, M., et~al.
\newblock Imagenet large scale visual recognition challenge.
\newblock \emph{International journal of computer vision}, 2015.

\bibitem[S{\o}nderby et~al.(2017)S{\o}nderby, Poole, and
  Mnih]{sonderby2017continuous}
S{\o}nderby, C.~K., Poole, B., and Mnih, A.
\newblock Continuous relaxation training of discrete latent variable image
  models.
\newblock In \emph{Beysian DeepLearning workshop, NIPS}, volume 201, 2017.

\bibitem[Takida et~al.(2022)Takida, Shibuya, Liao, Lai, Ohmura, Uesaka, Murata,
  Takahashi, Kumakura, and Mitsufuji]{takida2022sq}
Takida, Y., Shibuya, T., Liao, W., Lai, C.-H., Ohmura, J., Uesaka, T., Murata,
  N., Takahashi, S., Kumakura, T., and Mitsufuji, Y.
\newblock Sq-vae: Variational bayes on discrete representation with
  self-annealed stochastic quantization.
\newblock \emph{arXiv preprint arXiv:2205.07547}, 2022.

\bibitem[Tian et~al.(2020)Tian, Krishnan, and Isola]{tian2020contrastive}
Tian, Y., Krishnan, D., and Isola, P.
\newblock Contrastive multiview coding.
\newblock In \emph{European conference on computer vision}, pp.\  776--794.
  Springer, 2020.

\bibitem[Van Den~Oord et~al.(2017)Van Den~Oord, Vinyals, et~al.]{van2017neural}
Van Den~Oord, A., Vinyals, O., et~al.
\newblock Neural discrete representation learning.
\newblock \emph{Advances in neural information processing systems}, 30, 2017.

\bibitem[Williams et~al.(2020)Williams, Ringer, Ash, MacLeod, Dougherty, and
  Hughes]{williams2020hierarchical}
Williams, W., Ringer, S., Ash, T., MacLeod, D., Dougherty, J., and Hughes, J.
\newblock Hierarchical quantized autoencoders.
\newblock \emph{Advances in Neural Information Processing Systems},
  33:\penalty0 4524--4535, 2020.

\bibitem[Yan et~al.(2021)Yan, Zhang, Abbeel, and Srinivas]{yan2021videogpt}
Yan, W., Zhang, Y., Abbeel, P., and Srinivas, A.
\newblock Videogpt: Video generation using vq-vae and transformers.
\newblock \emph{arXiv preprint arXiv:2104.10157}, 2021.

\bibitem[Yu et~al.(2022)Yu, Li, Koh, Zhang, Pang, Qin, Ku, Xu, Baldridge, and
  Wu]{yu2021vector}
Yu, J., Li, X., Koh, J.~Y., Zhang, H., Pang, R., Qin, J., Ku, A., Xu, Y.,
  Baldridge, J., and Wu, Y.
\newblock Vector-quantized image modeling with improved vqgan.
\newblock In \emph{International Conference on Learning Representations}, 2022.

\bibitem[Zeghidour et~al.(2021)Zeghidour, Luebs, Omran, Skoglund, and
  Tagliasacchi]{zeghidour2021soundstream}
Zeghidour, N., Luebs, A., Omran, A., Skoglund, J., and Tagliasacchi, M.
\newblock Soundstream: An end-to-end neural audio codec.
\newblock \emph{IEEE/ACM Transactions on Audio, Speech, and Language
  Processing}, 30:\penalty0 495--507, 2021.

\bibitem[Zhang et~al.(2018)Zhang, Isola, Efros, Shechtman, and
  Wang]{zhang2018perceptual}
Zhang, R., Isola, P., Efros, A.~A., Shechtman, E., and Wang, O.
\newblock The unreasonable effectiveness of deep features as a perceptual
  metric.
\newblock In \emph{CVPR}, 2018.

\end{thebibliography}

\newpage
\clearpage

\appendix
\section{Appendix}
\subsection{Training details}
\label{app:train-details}

\xpar{VQ configuration and implementation}
We use $1024$ codes for all our experiments. $1024\sim4096$ is the typical codebook size used in prior works~\cite{esser2020taming,yan2021videogpt}. Using $4096$ codes does improve the performance slightly. We do not apply weight decay on the codebooks. For VQ hyper-parameters, we use $\alpha=5$ and $\beta=0.9\sim0.995$ with mean-squared error for the commitment loss. The performance starts to degrade outside this range of $\beta$. Note that using higher $\beta=1.0$ is equivalent to the EMA update. We implemented our own VQ algorithm to improve the run-time and memory efficiency. Standard implementation does not fit in memory for ImageNet training on standard commercial GPUs as the number of vectors grows close to 1 million for a given mini-batch. To mitigate this, we implemented our VQ algorithm using divide-and-conquer, which runs significantly faster and is more memory efficient compared to the naive implementation that allocates a contiguous memory to compute pair-wise distance. At a high level, our algorithm divides the batch of vectors into smaller chunks. For each chunk, batch matrix multiply ($\mathsf{cdist}$ in PyTorch) is applied for each subset and top-K reduction operation simultaneously. This is implemented in Python, and there is room for additional speed-up by implementing it in CUDA and parallelization via vmap. To further reduce computational and memory footprint, the distance is computed in half-precision, and the resulting code index is used to query the vector from float precision. Doing so adds no numerical imprecision to the existing model. 

For affine-reparameterization, we use the variant with learnable affine parameters, which we find to be easier to implement and more stable in practice. We find that the optimal learning rate scale for the affine parameters needs to be tuned based on the model architecture as the norm of the magnitudes grows differently for each model during training. For learnable affine parameters, this can be easily implemented in Python via:
\begin{align*}
&\mathtt{scale = 1 + aff\_lr\_scale * codebook\_scale}  \\
&\mathtt{bias = aff\_lr\_scale * codebook\_bias } \\
&\mathtt{codebook = scale * codebook + bias }
\end{align*}
To be robust to the affine-parameter learning rate scale, we recommend using norm constraint (e.g., max norm constraint, norm clipping) or placing the VQ layer after an explicit normalization layer. 

\xpar{Classification} 
For AlexNet and ResNet18, we follow the design choices in \url{https://github.com/pytorch/examples/blob/main/imagenet}. The training configuration can be found in~\tbl{table:alexnet-cfg} and~\tbl{table:resnet18-cfg}. For ViT(T), we started with the hyper-parameters recommended by~\cite{he2022masked} and re-tuned the hyper-parameters on the baseline VQ model. For ViT model codebase, we use the official PyTorch TorchiVison repository~\url{https://github.com/pytorch/vision/blob/main/torchvision/models/vision_transformer.py}. The ViT(T) configuration is from~\url{https://github.com/rwightman/pytorch-image-models/blob/main/timm/models/vision_transformer.py}. The architecture configuration is shown in~\tbl{table:vitt-arch} and the training configuration is shown in~\tbl{table:vitt-cfg}. ViT on ImageNet100 performs worse than ResNet18 as ViT does not perform very well on small datasets. This is a well-known observation~\cite{chen2021vision}. We apply data augmentation to the original image resolution and resize them to $224\times224$ for training.

\xpar{AlexNet quantization}
For AlexNet, we quantize the features after the convolutional layers and before the fully-connected layers.

\definecolor{Gray}{gray}{0.9}
\begin{table}[h!]
\begin{center}
\scalebox{0.85}{
\begin{tabular}{l|l} 
    \toprule
    config & value \\
    \midrule
    optimizer                   & SGD with momentum \\
    base learning rate          & $0.01$ \\
    weight decay                & 1e-4 \\
    optimizer momentum          & $0.9$ \\
    training epochs             & $90$   \\
    learning rate scheduler     & step \\
    step epochs                 & $30, 60$ \\
    augmentations               & RandomResizedCrop \\
    $\mathcal{L}_{\mathsf{cmt}}$ weight $\alpha$   & 5 \\
    $\mathcal{L}_{\mathsf{cmt}}$ tradeoff $\beta$  & 0.95 \\
    sync $\nu$                  & 2 \\
    affine lr scale             & 10 \\
\bottomrule
\end{tabular}
}
\caption{\small AlexNet image classification training configuration.}
\label{table:alexnet-cfg}
\end{center}
\end{table}

\xpar{ResNet18 quantization}
For ResNet18, we quantize after the second macroblock. This is after \texttt{layer2} in the TorchVision repository. This is roughly the halfway point in the ResNet18. We found this model architecture to be insensitive $\nu$, possibly due to batch-normalization.

\definecolor{Gray}{gray}{0.9}
\begin{table}[h!]
\begin{center}
\scalebox{0.85}{
\begin{tabular}{l|l} 
    \toprule
    config & value \\
    \midrule
    optimizer                   & SGD\\
    base learning rate          & $0.1$ \\
    weight decay                & 1e-4 \\
    optimizer momentum          & $0.9$ \\
    training epochs             & $90$   \\
    learning rate scheduler     & step \\
    step epochs                 & $30, 60$ \\
    augmentations               & RandomResizedCrop \\
    $\mathcal{L}_{\mathsf{cmt}}$ weight $\alpha$   & 5 \\
    $\mathcal{L}_{\mathsf{cmt}}$ tradeoff $\beta$  & 0.98 \\
    sync $\nu$                  & 0.2 \\
    affine lr scale                   & 1 \\
\bottomrule
\end{tabular}
}
\caption{\small ResNet18 image classification training configuration.}
\label{table:resnet18-cfg}
\end{center}
\end{table}

\xpar{ViT quanitzation}
ViT does not directly operate on image pixels but on image patches. Hence, the quantization cannot be applied to pixels. For our experimental setup, we tokenize non-overlapping patches of size $16\times16$ resulting in a total of $14\times14=196$ input tokens. These individual tokens are quantized. Note that this is often much fewer than the number of embedding vectors used in CNNs (\eg feature embedding of $32\times32=1024$ embedding vectors) and may be the reason why they suffer more from vector-quantization. We apply the quantization after the 6th transformer block. Training ViT without replacement policy is extremely finicky, which requires careful hyper-parameter tuning. When using replacmenet policy, it becomes more robust to wide-range of hyper-parameters. We recommend re-tuning the configuration when using it jointly with replacement policy.

\definecolor{Gray}{gray}{0.9}
\begin{table}[ht!]
\begin{center}
\scalebox{0.85}{
\begin{tabular}{l|l} 
    \toprule
    config & value \\
    \midrule
    optimizer                   & AdamW\\
    base learning rate          & 2e-4 \\
    weight decay                & $0.03$ \\
    optimizer momentum          & $(0.9, 0.95)$ \\
    training epochs             & $90$   \\
    learning rate scheduler     & Cosine \\
    warmup epochs               & $10$ \\
    augmentations               & RandomResizedCrop \\
   $\mathcal{L}_{\mathsf{cmt}}$ weight $\alpha$   & 5 \\
    $\mathcal{L}_{\mathsf{cmt}}$ tradeoff $\beta$  & 0.995 \\
    sync $\nu$                  & 0.01 \\
    affine lr scale              & 10 \\
\bottomrule
\end{tabular}
}
\caption{\small ViT-Tiny image classification training configuration.}
\label{table:vitt-cfg}
\end{center}
\end{table}

\definecolor{Gray}{gray}{0.9}
\begin{table}[h!]
\begin{center}
\scalebox{0.85}{
\begin{tabular}{l|l} 
    \toprule
    config & value \\
    \midrule
    patch size       & $16 \times 16$\\
    embedding dim    & 192 \\
    depth            & 12 \\
    num heads        & 3 \\
    feedforward dim  & 768 \\
    activation       & GELU \\
\bottomrule
\end{tabular}
}
\caption{\small ViT-Tiny model architecture.}
\label{table:vitt-arch}
\end{center}
\end{table}

\xpar{Generative modeling}
For generative modeling, we use backbone architecture from~\cite{takida2022sq} with $64$ channels. The training configuration is listed in~\tbl{table:gen-cfg}. We use MSE for reconstruction loss, and we do not use any perceptual or discriminative loss. For CIFAR10, we use an image size of $32\times32$, and for CelebA, we use an image size of $128\times128$.

For MaskGIT~\cite{chang2022maskgit}, we followed the author's original codebase and reimplemented it in PyTorch. To ensure that we can fit the model in a commercial GPU ($24$GB), we reduced the number of channels by 1/4th ($32$ channels) for the auto-encoder and used an $8$-layer transformer instead of $24$. The code resolution factor is $8$, resulting in an $8\times8$ code map. A transformer is trained on the vectorized code. See~\cite{chang2022maskgit} for the method details.

\definecolor{Gray}{gray}{0.9}
\begin{table}[h!]
\begin{center}
\scalebox{0.85}{
\begin{tabular}{l|l} 
    \toprule
    config & value \\
    \midrule
    optimizer                   & AdamW\\
    base learning rate          & 1e-4 \\
    weight decay                & 1e-4 \\
    optimizer momentum          & $(0.9, 0.95)$ \\
    training epochs             & $90$   \\
    learning rate scheduler     & Cosine \\
    warmup epochs               & $10$ \\
    augmentations               & None \\
\bottomrule
\end{tabular}
}
\caption{\small Generative modeling, auto-encoder configuration for CIFAR10 and CelebA.}
\label{table:gen-cfg}
\end{center}
\end{table}

\xpar{Baseline}
For \textbf{SQVAE}~\cite{takida2022sq}, we use the Gaussian-SQVAE (4th-variant), which is the best-performing variant for unconditional generative modeling that uses a diagonal covariance matrix. We used the architecture from the official codebase and replicated their stochastic quantization layer to interface our framework. We anneal the temperature till convergence with a geometric decay. We re-tuned the learning rate and the weighting of the loss. We find the default scaling of $1\sim0.1$ to work, setting it any higher did not train. SQVAE has an entropy term that encourages diversity in the codebook along with the ELBO objective. We observed that the model can easily collapse if the temperature is annealed too fast and, therefore, requires much longer to converge.

\textbf{Gumbel-VQ} was a variation proposed in the public repository by~\cite{dvq2021karpathygithub} and was also been used by~\cite{esser2020taming} in their codebase. Similar to~\cite{takida2022sq}, Gumbel-VQ minimizes the ELBO; however, unlike standard VQ methods, which compute the distance across all code vectors, Gumbel-VQ predicts a distribution over the code without making any explicit comparisons. The model then uses the Gumbel-softmax~\cite{jang2016categorical} trick to sample from the distribution. We tried various hyper-parameters for the ELBO loss weight \{5e-1, 5e-2, 5e-3, 5e-4, 5e-5\} and found 5e-3 to work the best.

\begin{figure*}[t!]
     \centering
     \includegraphics[width=0.9\linewidth]{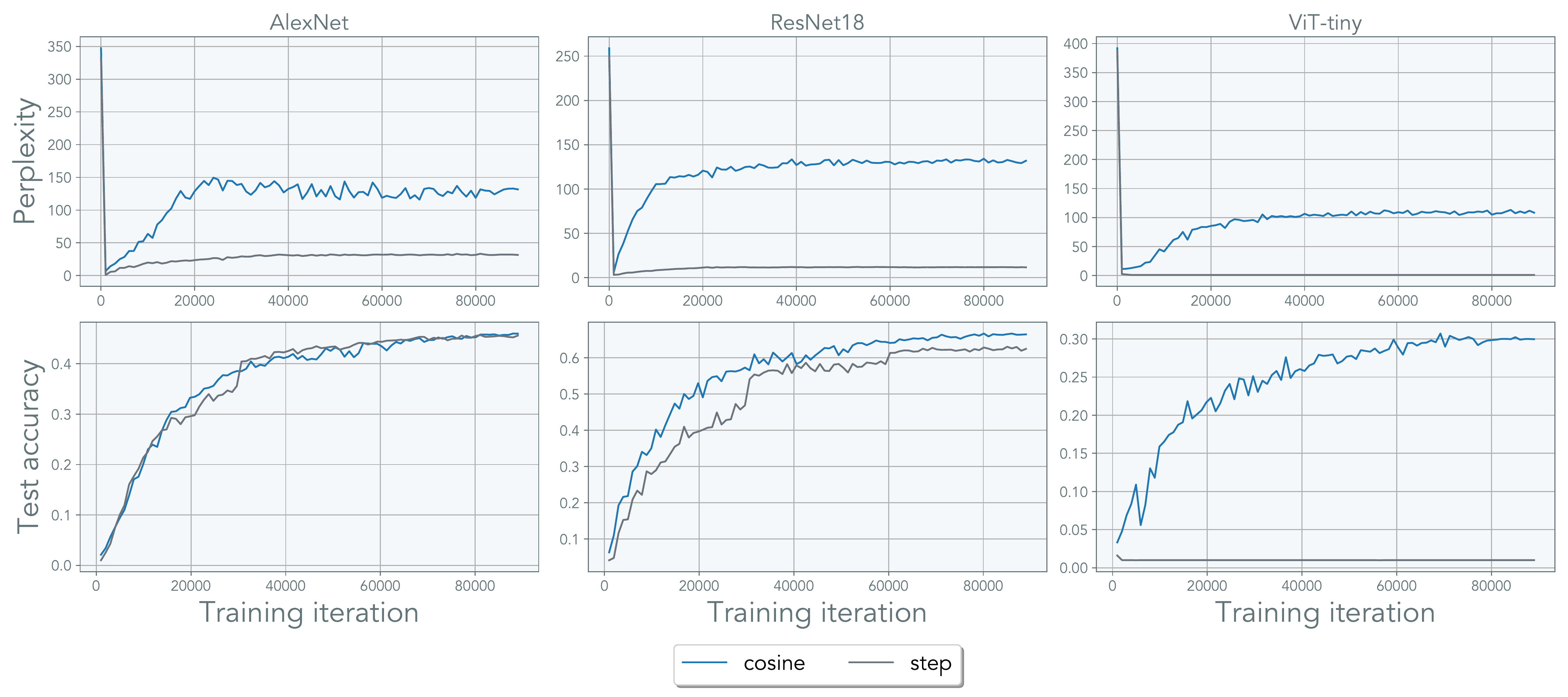}
     \caption{\small\textbf{Warmup improves perplexity:} We plot the perplexity and test accuracy for methods trained with and without a linear warmup. All methods were initialized with K-means.}
     \label{fig:cos}
\end{figure*}

\subsection{EMA and commitment loss}
\label{app:ema}
We show the equivalence between EMA and the commitment loss. Let $\beta=1$ for the commitment loss with MSE for $d$:

\vspace{-0.2in}
\begin{align}
\mathcal{L}_{\mathsf{vq}}(\bz_e^{(t)}, \bz_q^{(t)}) = \frac{1}{2}\lVert \bz_e^{(t)} - \bz_q^{(t)} \rVert_2^2
\end{align}
\vspace{-0.2in}

The gradient of the commitment loss is computed with respect to the codes $\bz_q$ is:

\vspace{-0.2in}
\begin{align}
\frac{\partial \mathcal{L}_{\mathsf{vq}}(\bz_e^{(t)}, \bz_q^{(t)})}{\partial \bz_q^{(t)}} &= \frac{1}{2}\lVert \bz_e^{(t)} - \bz_q^{(t)} \rVert_2^2 \\
&= \bz_e^{(t)} - \bz_q^{(t)} \\
\end{align}
\vspace{-0.2in}

Then the update for $\bz_q^{(t+1)}$ is:

\vspace{-0.2in}
\begin{align}
\bz_q^{(t+1)} &= \bz_q^{(t)} - \eta \frac{\partial \mathcal{L}_{\mathsf{vq}}(\bz_e^{(t)}, \bz_q^{(t)})}{\partial \bz_q^{(t)}} \\
&= \bz_q^{(t)} - \eta \cdot (\bz_e^{(t)} - \bz_q^{(t)}) \\
&= (1 - \eta) \cdot \bz_e^{(t)} + \eta \cdot \bz_q^{(t)}
\end{align}
\vspace{-0.2in}

Letting $\eta=\gamma$ and using SGD to optimize the code-vectors, we recover the EMA update rule in~\eqn{eqn:ema}

\begin{figure}[t!]
     \centering
     \includegraphics[width=0.9\linewidth]{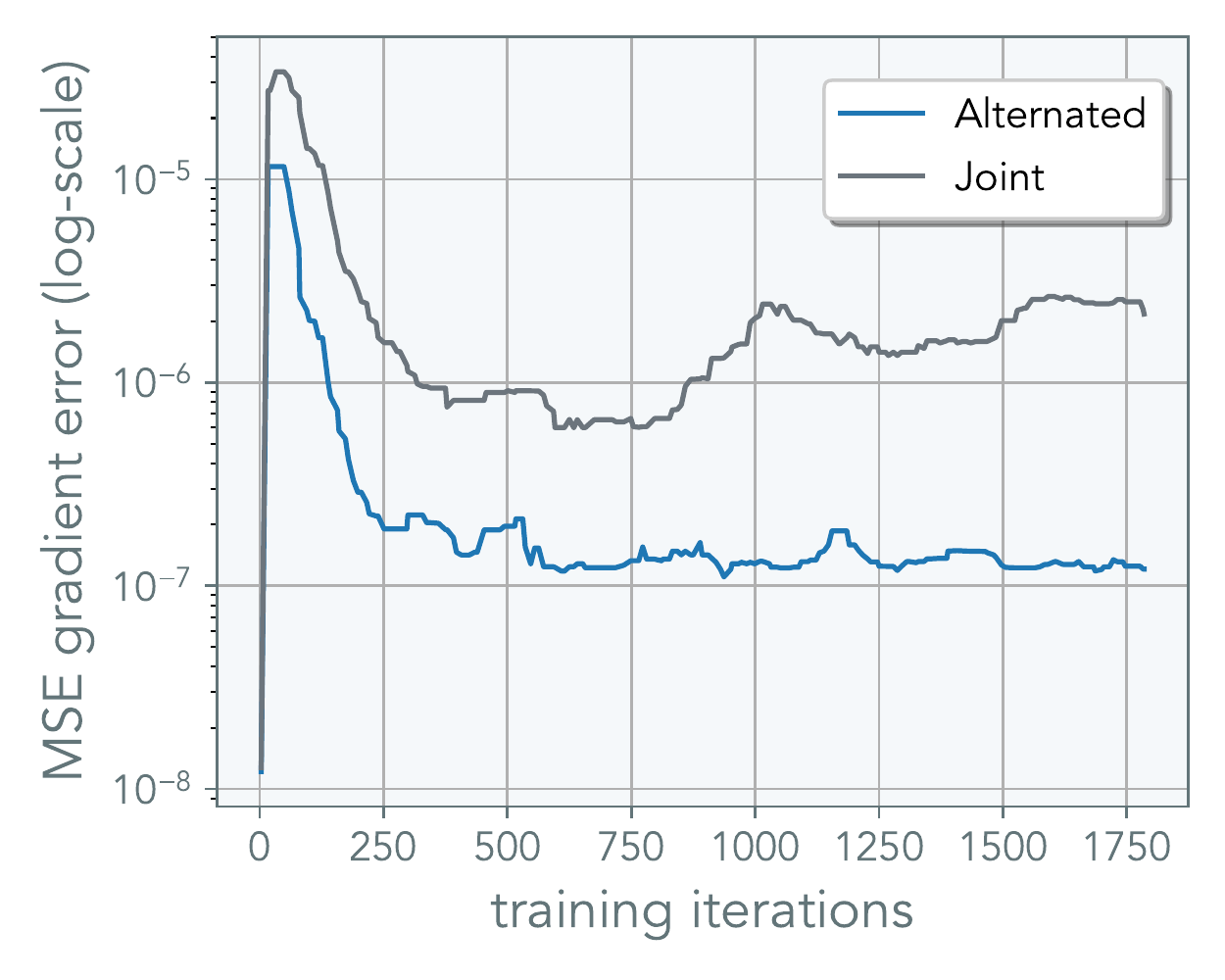}
     \caption{\small\textbf{Gradient estimation gap:} STE estimation gap comparing standard joint optimization versus alternated optimization. The y-axis is log-scale.}
     \label{fig:grad-err}
\end{figure}
\begin{figure}[t!]
     \centering
     \includegraphics[width=1.0\linewidth]{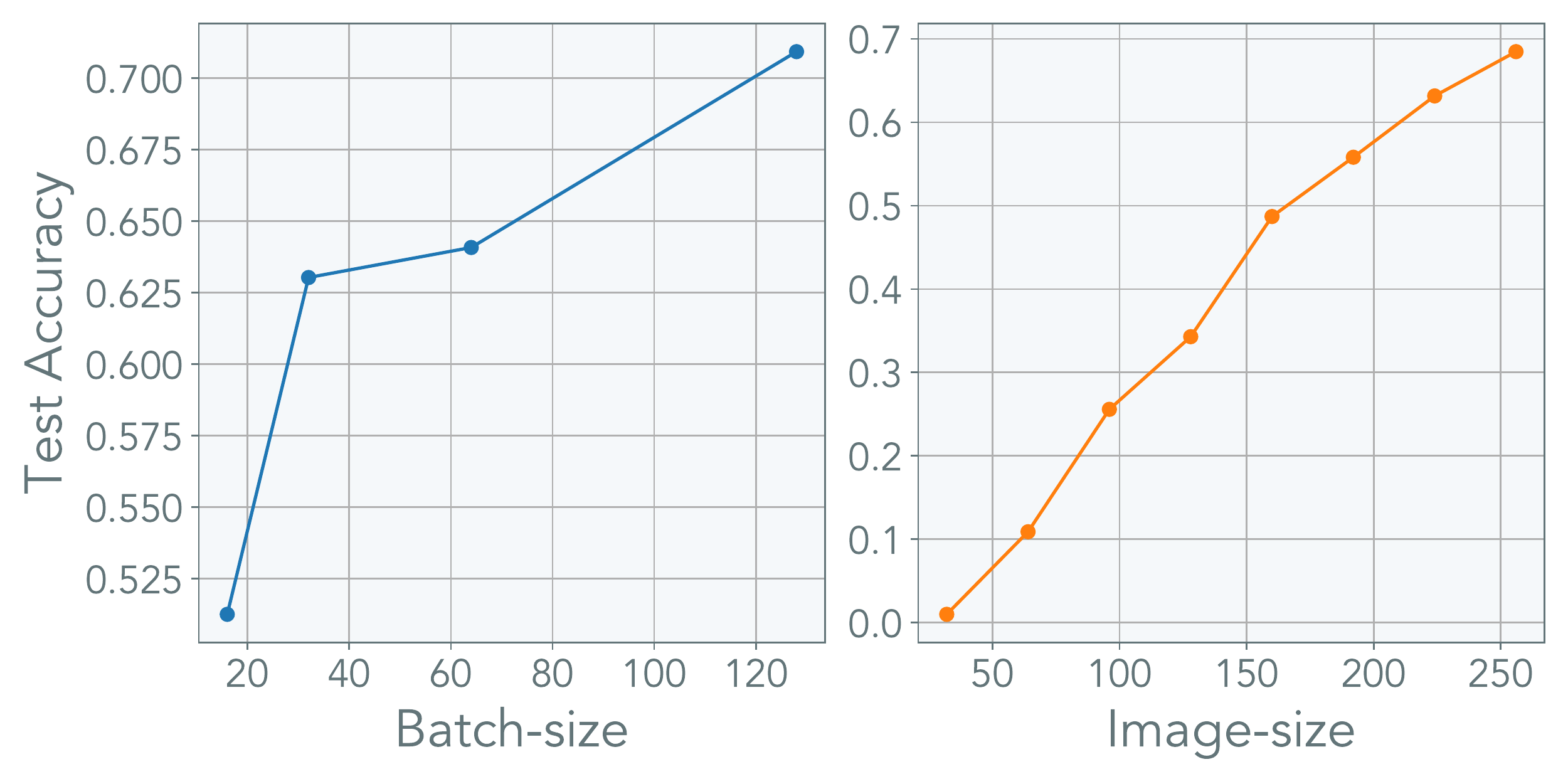}
     \caption{
     \small
     \textbf{How activation rate affects performance:} 
     The performance of VQNs is tied to the selection likelihood of the code-vectors $p_{\mathsf{activate}}$. Here we visualize how the batch-size and the image-size affects the model performance. VQNs that have a low activation ratio perform signifcantly worse than standard networks. On the right, we convert the x-axis into $p_{\mathsf{activate}}$ using~\eqn{eqn:activate}}.
     \label{fig:activation}
\end{figure}

\begin{figure*}[!htb]
     \centering
     \includegraphics[width=0.49\linewidth]{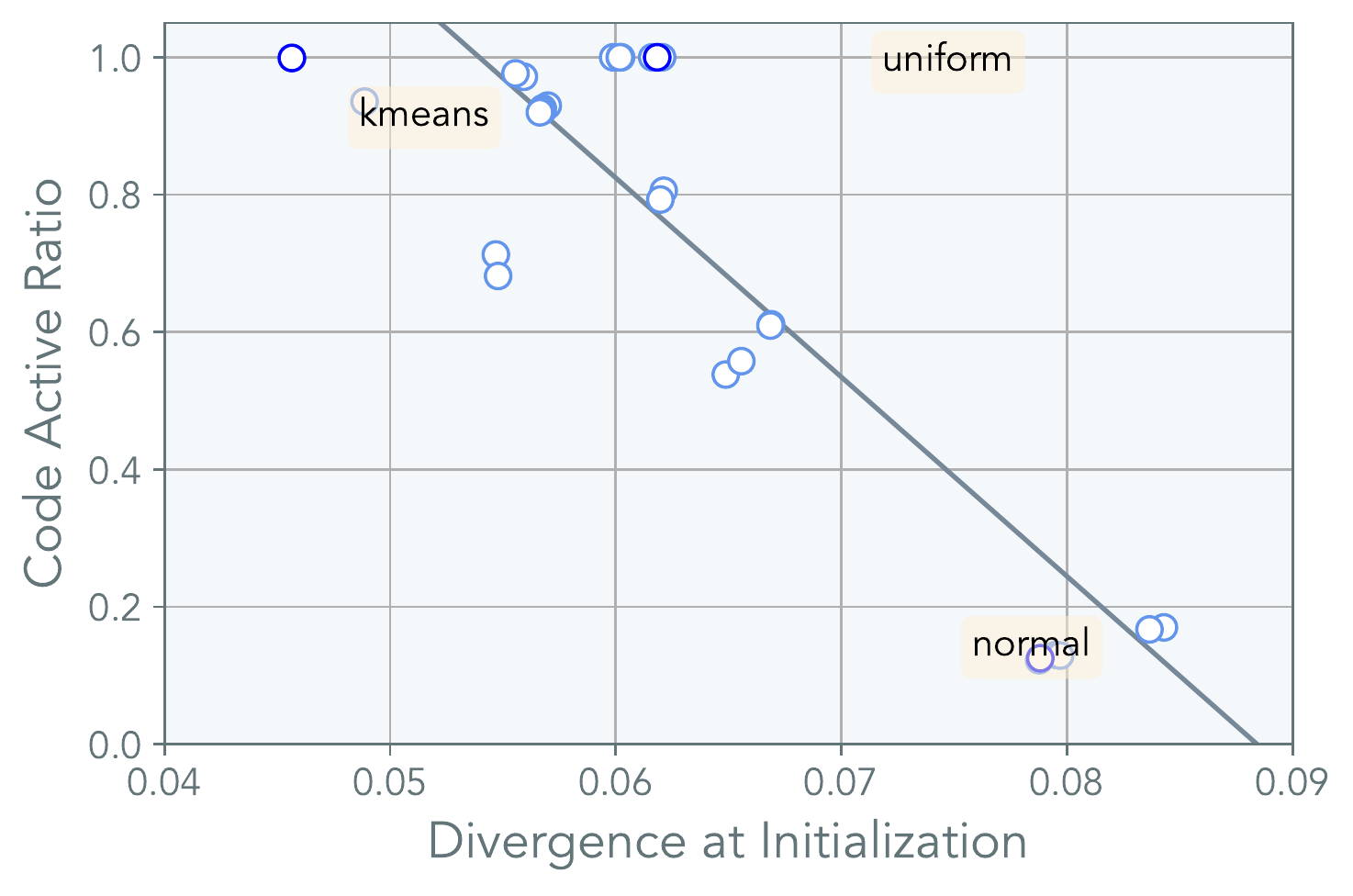}
     \includegraphics[width=0.49\linewidth]{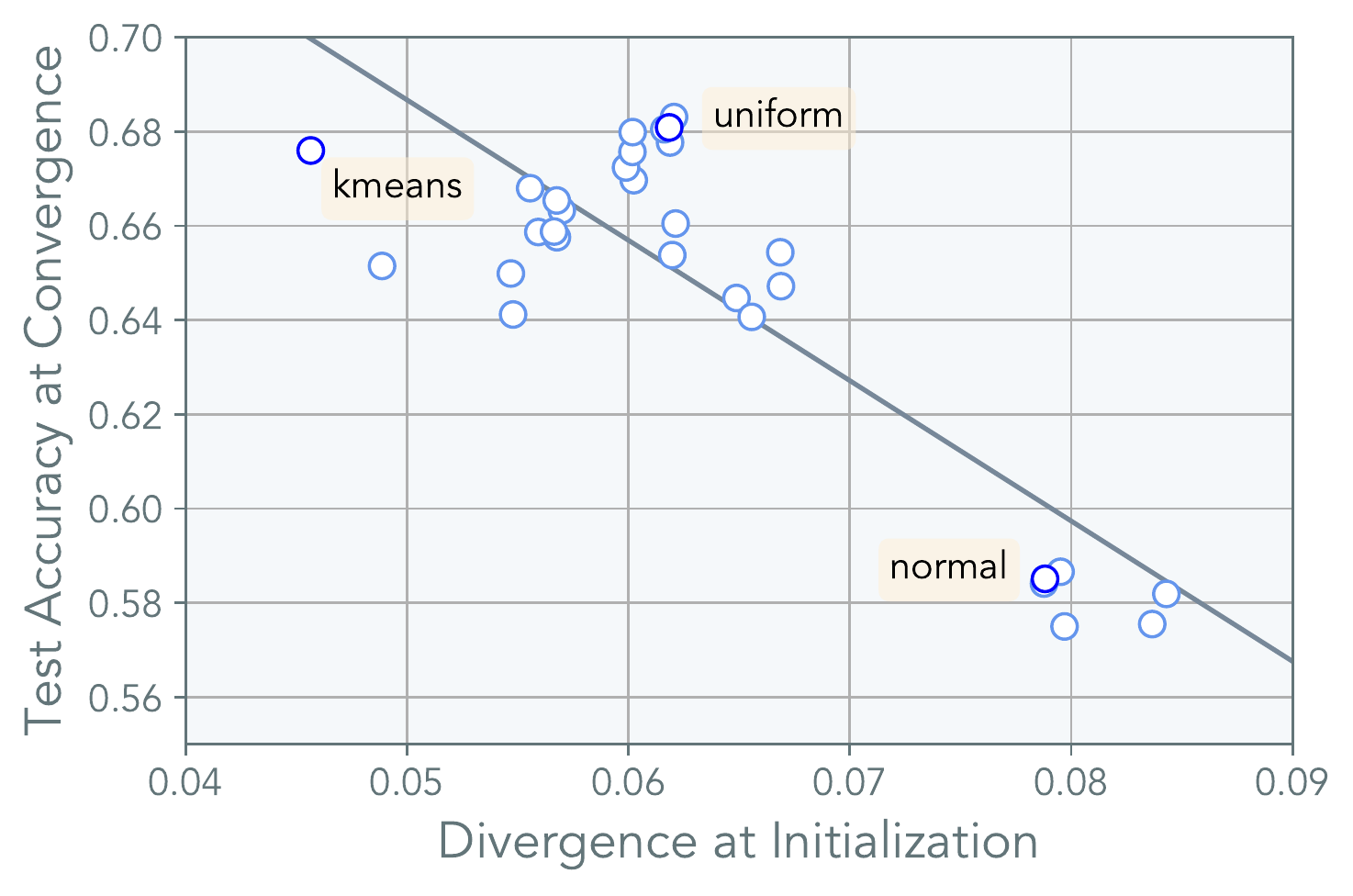}
     \caption{
     \small
     \textbf{Divergence vs Accuracy:} 
     We visualize how the divergence between $\mathcal{P}_z$ and $\mathcal{Q}_z$ affect the performance of the model at convergence. We use various standard initialization schemes and provide some labels (the full list is reported in the appendix). We observed a linear relationship between the divergence at initialization and performance at convergence. We found that the quantization error at initialization determines the effect of codebook collapse.
     }
     \label{fig:div-init}
\end{figure*}

\begin{figure}[t!]
     \centering
     \includegraphics[width=1.0\linewidth]{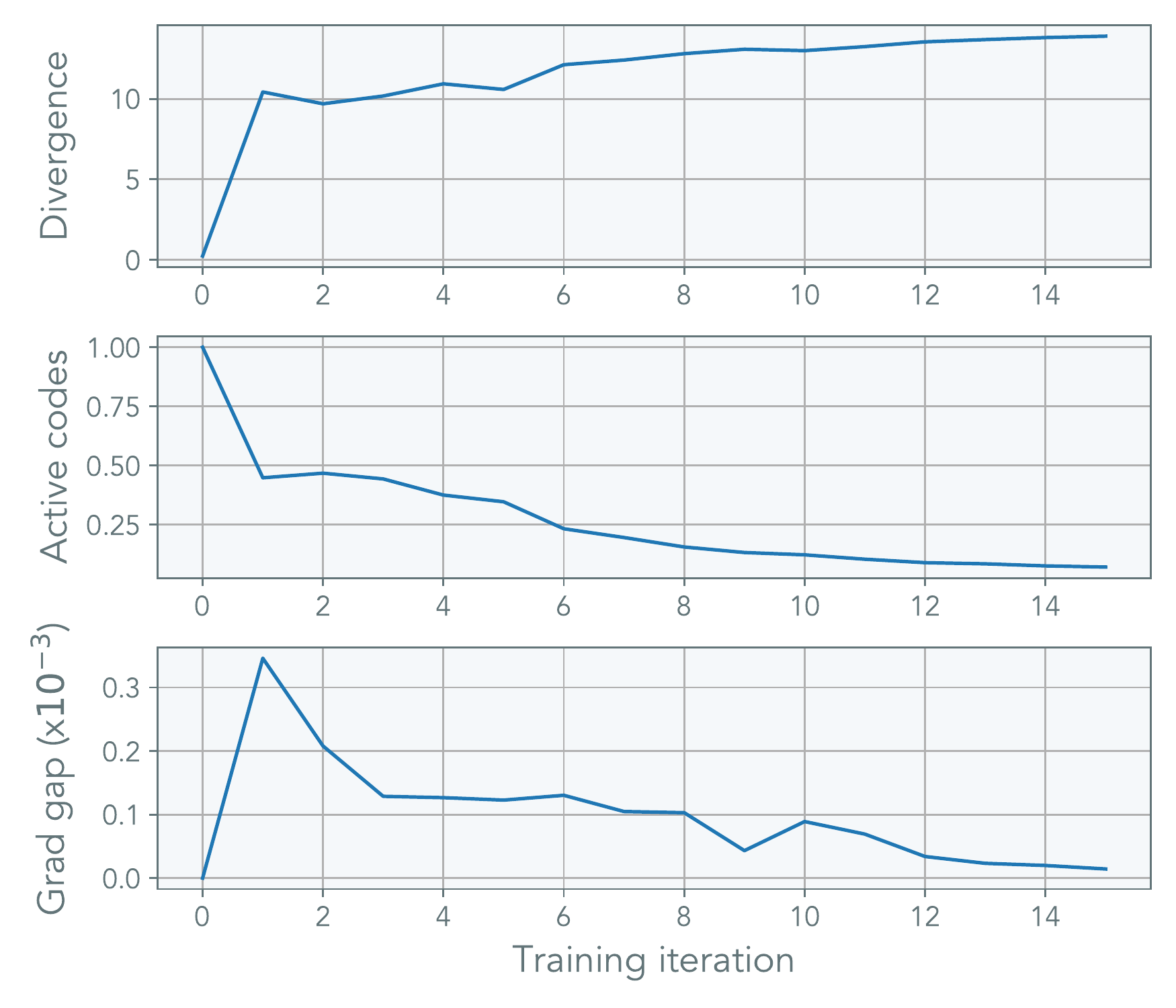}
     \caption{
     \small
     \textbf{Codebook collapse early on in training:} 
     The dynamics of vector-quantized ResNet18 in the first $16$ iteration of training. The divergence code is measured between $D(\mathcal{C}_{\bz}, \mathcal{Q}_{\bz})$. The model is initialized using K-means. Despite having ideal conditions, the model starts to degrade after a single iteration of SGD update.} 
     \label{fig:early-collapse}
\end{figure}

\subsection{Gradient estimation gap}
\label{app:grad-err}
To compute the gradient estimation gap in~\eqn{eqn:grad-err}, we compute $2$ forward passes. Once with the quantization function and once without. Let $g^{(l)}$ be the gradient without the quantization function for layer $l \in L$ and $\hat{g}^{(l)}$ be the gradient gap computed without the quantization error. Then the total average gradient error is $\sum_{l \in L} \lVert g^{(l)} - \hat{g}^{(l)} \rVert^2_2$. We visualize the gradient error of VQVAE training for the baseline model and our model that uses alternated optimization~\fig{fig:grad-err}. The VQ-layers use $l_2$ normalization, and the gradient gap is computed in log-scale -- hence the gradient gap is much larger than it appears in the figure.

\subsection{Warmup improves perplexity and performance}
\label{app:warmup}
As mentioned in~\sect{sec:warmup}, we found using a warmup to significantly improve codebook perplexity. This, in turn, results in better performance. In~\fig{fig:cos}, we compare VQNs trained with and without warmup. All methods are initialized using K-means. The baseline model uses a step scheduler. For ViT, the model collapses when we do not use a linear warmup. We hypothesize that a small learning rate allows the code-vectors to catch up to the model representation, while using a large learning rate at initialization causes misalignment between code-vectors and embedding vectors, resulting in index collapse.

\subsection{Sensitivity of VQ models}
\label{app:sensitivity}
Vector-quantization using the standard commitment loss result in sparse gradients by design. Therefore, it is imperative to isolate design factors that would contribute to this sparsity. Since sparsity is directly correlated with the selection rate of the code vectors, we can write out the selection probability under a simplifying assumption of i.i.d selection rate. For a VQN that operates on images, the likelihood that a code-vector $\mathbf{c}_i$ will activate at-least $k$ times is:

\vspace{-0.15in}
{
\small
\begin{align}
p&(\mathbf{c}_i \;\text{activate atleast}\; k \; \text{times}) = \nonumber\\
& 1 - \sum_{j=0}^{k-1} {bhw/ 2^{n_{\mathsf{pool}}} \choose j} \left(\frac{1}{\mathcal{C}}\right)^j \left(1 - \frac{1}{\mathcal{C}}\right)^{bhw/ 2^{n_{\mathsf{pool}}} - j}
\end{align}
\label{eqn:activate}
}
\vspace{-0.15in}

Where $h$ and $w$ are the image dimensions and $n_{\mathsf{pool}}$ is the number of $2\times2$ pooling layers, and $b$ is the batch size. The equation above is simply a summation of the binomial distribution over the selection probability. Here it becomes apparent that the image size, the batch size, the codebook size, and the number of pooling layers all directly affect the selection likelihood. While it is impossible to study the effect of these factors in perfect isolation, in~\fig{fig:activation}, we show how performance degrades compared to the non-quantized network. %

\subsection{Index Collapse}
\label{app:early-collapse}
Index collapse is a phenomenon associated with the under-utilization of the codebook. While code-vectors become inactive throughout training, it is common to see a sudden collapse in code-vector usage early in training. In~\fig{fig:early-collapse}, we visualize the index collapse on ResNet18 trained with K-means initialization. At initialization, all codes are actively used, with the gradient gap and divergence being close to 0. Here the divergence is measured between $D(\mathcal{C}_{\bz}, \mathcal{Q}_{\bz})$, which measures the bifurcation of the codebook distribution. After a single iteration of SGD update, the code-vector starts to diverge, with the number of active codes dropping below $50$\%. The resulting misalignment causes a large spike in the gradient error gap. The divergence continues to grow with the number of active codes approaching $1\%$ utilization. With fewer active codes, the encoder learns a degenerate solution of predicting these few remaining code-vectors.

\begin{figure*}[t!]
    \centering
    \includegraphics[width=0.33\textwidth]{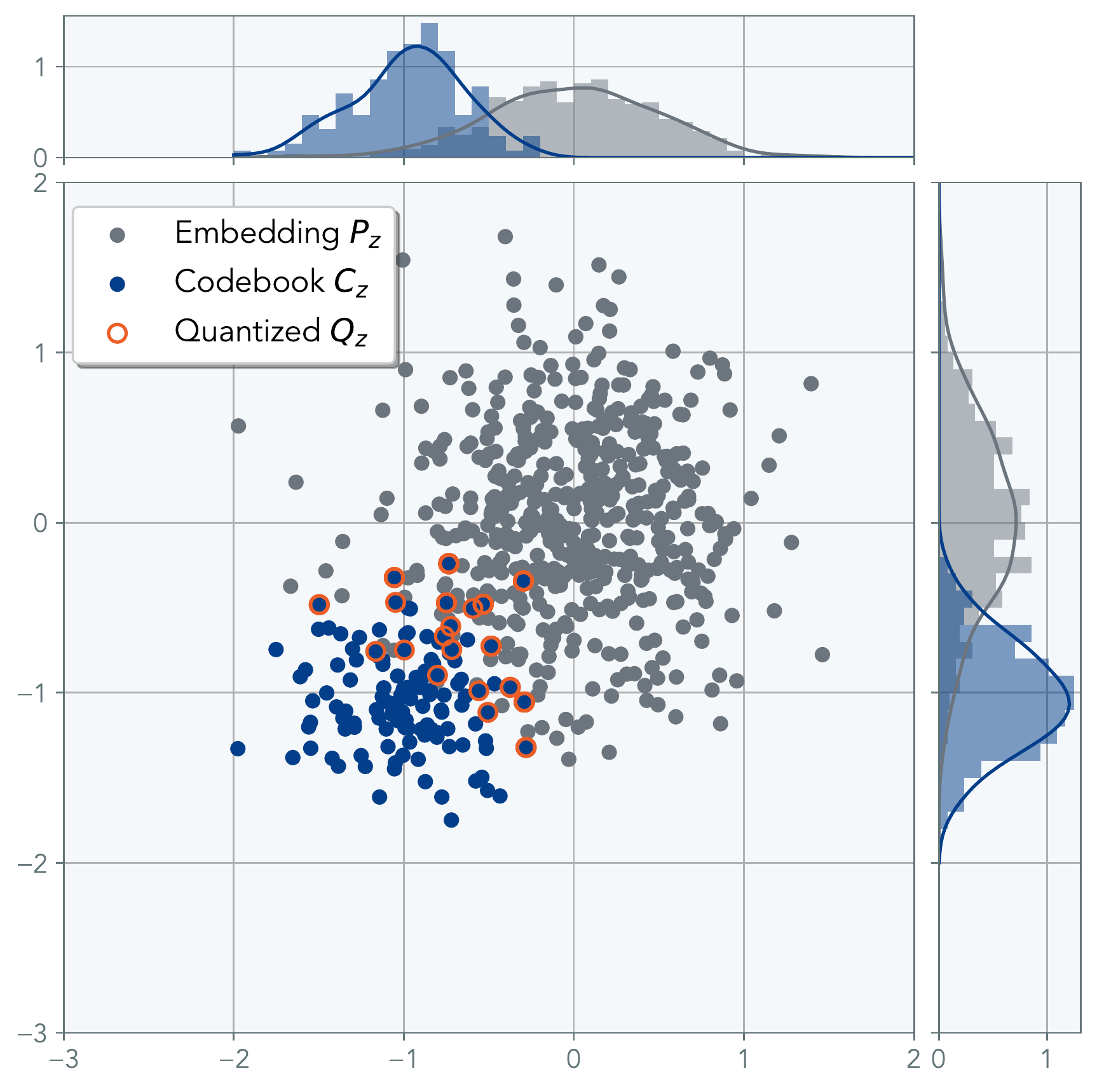}
    \hfill
    \includegraphics[width=0.33\textwidth]{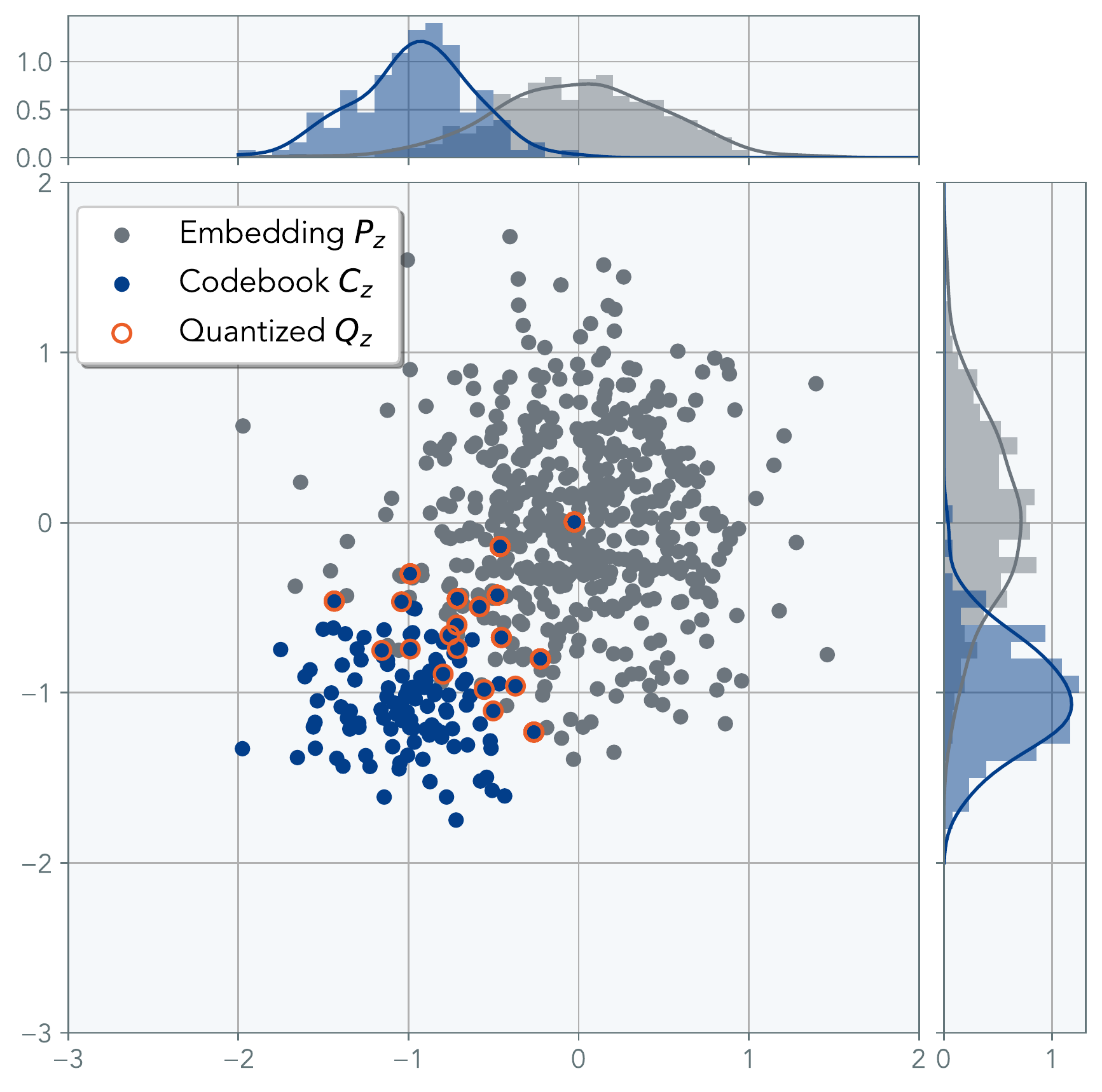}
    \hfill
    \includegraphics[width=0.33\textwidth]{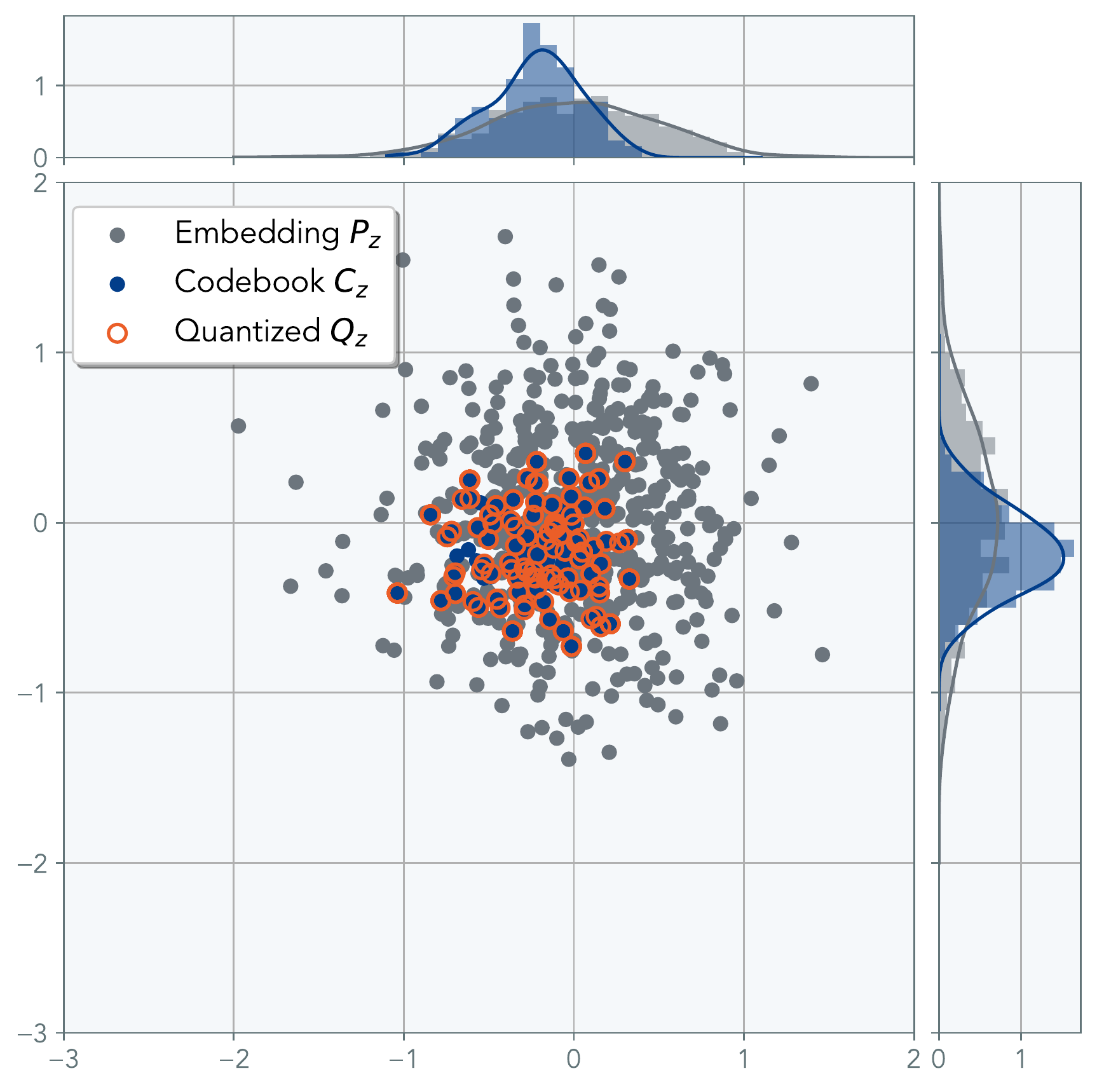}
    \caption{
    \small
    \textbf{Minimizing internal codebook covariate shift with affine parameterization:} 
    Consider a toy example in 2D with the embedding distribution $\mathcal{P}_{\bz} \sim \mathcal{N}([0, 0]^{\mathsf{T}}, 0.5 \cdot \mathbf{I})$ in grey and codebook distribution $\mathcal{C}_{\bz} \sim \mathcal{N}([-1, -1]^{\mathsf{T}}, 0.3 \cdot \mathbf{I})$ in blue, and the selected quantized points $\mathcal{Q}_{\bz}$ highlighted in orange. Here $|\mathcal{P}_{\bz}| = 512$ and $|\mathcal{Q}_{\bz}| = 128$. We visualize the distribution along with the density after $20$ codebook updates using the standard commitment loss with $\beta=0$ (same as the EMA-update variant) and a learning rate of $0.1$. During initialization, only few samples are selected by the embedding distribution. Therefore, when using standard parameterization, more than $90\%$ of the codes are not updated. For affine-parameterization, we consider a learnable variant. When using affine-parameterization, even if the codes are not selected, all the codes receive gradients and are able to better match the embedding distribution. 
    }
    \label{fig:affine}
\end{figure*}

\subsection{The effect of VQ initialization}
\label{app:init}
To improve gradient estimates, one can ensure the quantization error is small. 
To do so, one can ensure that the quantization error is small at initialization by choosing an appropriate weight initialization scheme. Unfortunately, Initializing the codebooks requires a priori knowledge of the model architecture and data distribution, as the distribution associated with the input embedding $\bz_e$ is hard to calculate ahead of time nor may not have a closed form distribution to sample from. For example, a VQ layer placed after a ReLU function implies that code-vectors in the negative half-space will never be sampled.
\definecolor{Gray}{gray}{0.9}
\newcolumntype{g}{>{\columncolor{Gray}}c}
\newcolumntype{P}[1]{>{\centering\arraybackslash}p{#1}}
\begin{table}[t!]
\begin{center}
\scalebox{0.8}{
\begin{tabular}{l|l|c} 
    \toprule
     & Method & FID $\downarrow$ \\
    \midrule
    {\multirow{4}{*}{\rotatebox[origin=c]{90}{$\mathsf{CELEBA}$}}}
    & MaskGIT*                & 90.4 \\  
    & + $l_2$                & 81.5 \\  
    & + replace              & 79.7 \\  
    & + Affine + OPT (ours)  & \textbf{74.8} \\  
    \bottomrule
\end{tabular}
}
\caption{\small \textbf{MaskGIT generation:} FID on CelebA image generation using MaskGIT. *We use a significantly smaller capacity architecture to make the training feasible.}
\label{table:maskgit}
\end{center}
\end{table}

To mitigate this issue, it is a common practice to use data-dependent initialization such as K-means. 
These methods precisely capture the distribution of $\bz_e$ and better distributes the likelihood individual codebooks will be sampled. 
In fig X, we show the relationship between quantization error at initialization and the final performance of the model.
Moreover, we find that the quantization error at initialization is a strong indicator for index-collapse / under-utilization of code-vectors -- a phenomenon in which the number of active code-vectors at convergence is significantly smaller than the one we started with. As shown in~\fig{fig:div-init}, a good initialization scheme mitigates index-collapse and leads to favorable performance.

\begin{figure*}[t!]
     \centering
     \includegraphics[width=0.495\linewidth]{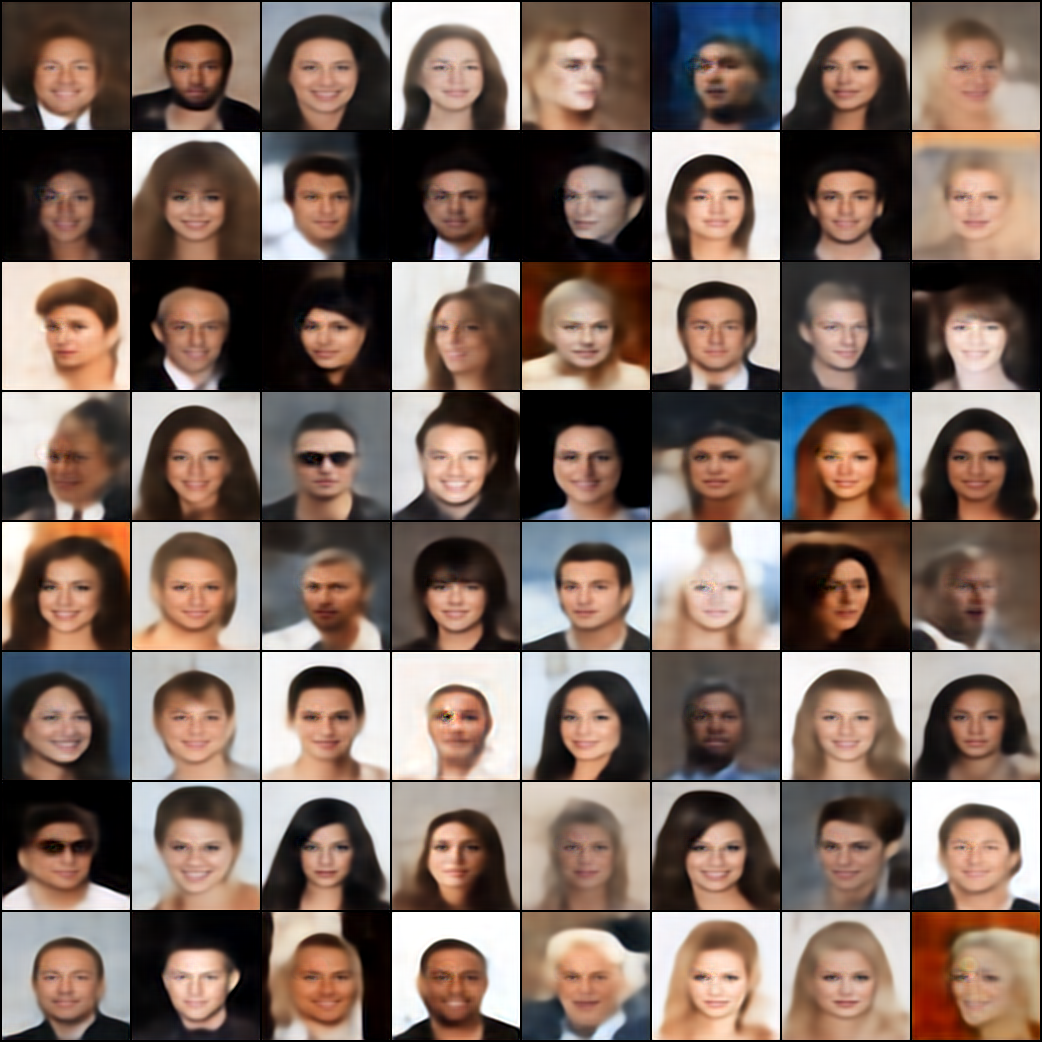}
     \hfill
    \includegraphics[width=0.495\linewidth]{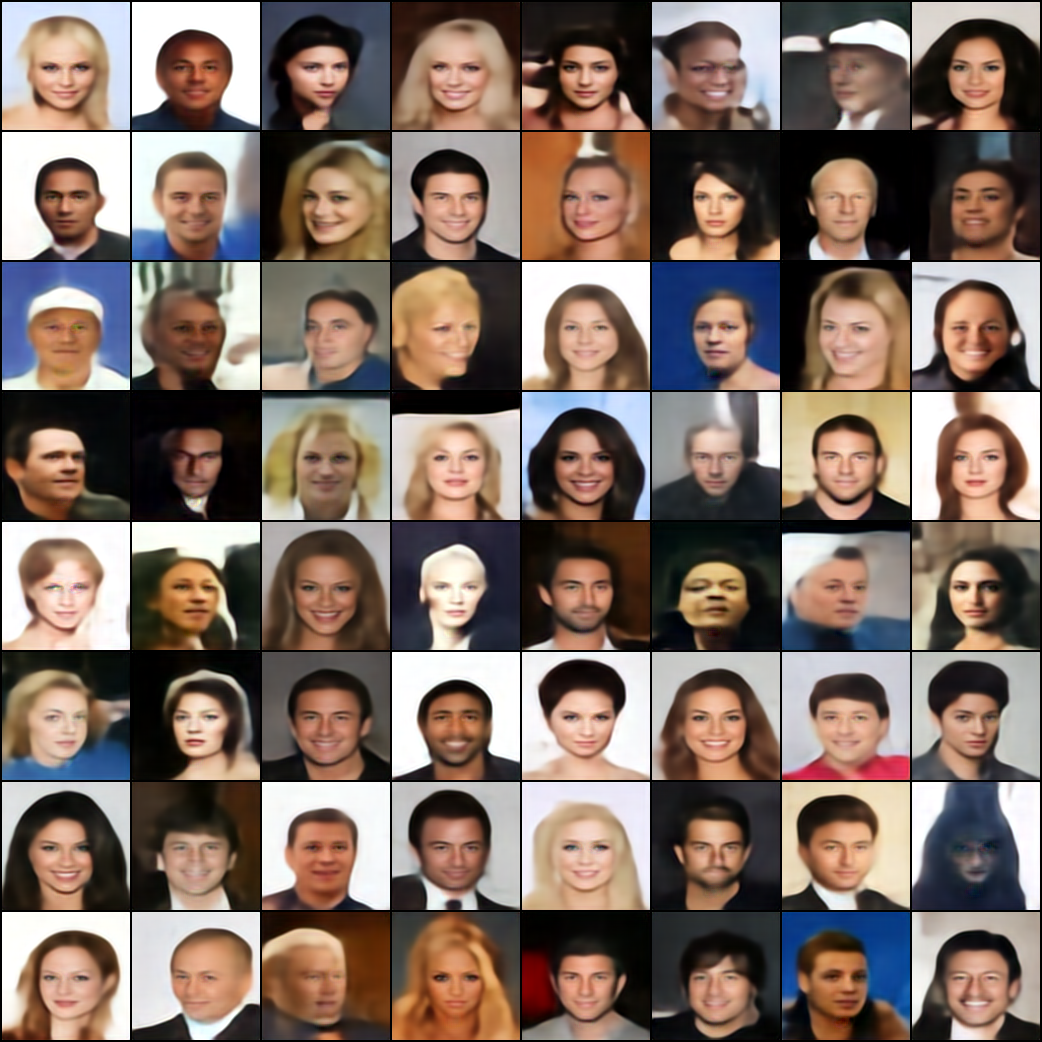}

     \caption{\small\textbf{MaskGIT on CelebA:} Generation result on $128\times128$ CelebA images. On the left we have the standard MaskGIT training, on the right we have MaskGIT with our proposed method.}
     \label{fig:maskgit}
\end{figure*}

\subsection{Generation with MaskGIT}
We extend our results on generative modeling using MaskGIT~\cite{chang2022maskgit} on CelebA. In~\tbl{table:maskgit}, the generation results improve from the baseline by $15.6$ FID and $4.9$ FID from the best-performing variation. We did not scale to images to compute FID. In~\fig{fig:maskgit}, we visualize randomly sampled images from the model. No samples were cherry sampled.

\subsection{Affine reparameterization on toy setting}
\label{app:affine-toy}
The use of vector quantization adds to an already existing issue of internal covariate shift. When the internal representation, $\mathcal{P}{\bz}$, is updated, it takes a longer time for the codebook, $\mathcal{C}{\bz}$, to catch up. Furthermore, if the update in $\mathcal{P}{\bz}$ is too large, for example, due to a high learning rate, the assignment of codes to embeddings can become severely misaligned. To visualize this, consider a toy example illustrated in~\fig{fig:affine} where the embedding distribution, $\mathcal{P}{\bz}$, and the codebook distribution, $\mathcal{C}_{\bz}$, undergo a drift (left). As a result, in the next iteration, only a small fraction of the codes are chosen and updated, while the rest remain unchanged. This misalignment in the codebook and embedding distributions leads to suboptimal model performance. Using affine reparameterization can mitigate this by allowing gradients to flow through all code-vectors implicitly via the shared parameters. The toy example uses the learnable parameter variant.

When implementing affine reparameterization, we observed better performance when using accumulated statistics. We accumulate the statistics with a momentum of $0.1$. We compute the moments of both $\bz_e$ and $\mathcal{C}$ and compute the appropriate shift the match the moments of $\bz_e$. We reduce the weighting of the commitment loss to $\alpha=2$. When using the learnable parameters variant, while it performs slightly worse, the implementation only requires a $2$ line change.

\subsection{Discussion and intuition of the commitment loss}
\label{app:cmt}
We discuss the difficulty of achieving perfect assignments using commitment loss. First, the upper bound of the commitment loss is given by:

\vspace{-0.15in}
\begin{align}
\mathcal{L}_{\mathsf{commit}} &=\frac{1}{2 \lvert \mathcal{P}_{\bz} \rvert} \sum_{\bz_i \sim \mathcal{P}_{\bz}} \min_{\bc_j \sim \mathcal{C}} \lVert \bz_i - \bc_j \rVert^2 \\
&\leq \min_{\bc_j \sim \mathcal{C}} \frac{1}{2 \lvert \mathcal{P}_{\bz} \rvert} \sum_{\bz_i \sim \mathcal{P}_{\bz}} \lVert \bz_i - \bc_j \rVert^2 \\
&= \frac{1}{2}\min_{\bc_j \sim \mathcal{C}} \mathbb{E}_{\mathcal{P}_{\bz}} \left[ \lVert \bz_i - \bc_j \rVert^2 \right] \\
& = \frac{1}{2} Var(\mathcal{P}_{\bz})
\end{align}
\vspace{-0.15in}

Where the minimum is achieved when $c_j = \mu({\mathcal{P}_{\bz}})$, the mean of the embedding. By moving the minimization function outside, we optimize with respect to a single code vector.

The tight lower bound of the commitment loss can be achieved by solving an assignment problem. Let $\mathcal{B} = \{ B_0, B_1, \dots, B_m\}$, where $|\mathcal{C}| = m$. Each cluster ball $B_i$ is a cluster associated with the code-vector $c_i$. Using this notation, we rewrite the commitment loss as:

\vspace{-0.15in}
\begin{align}
&\argmin_{\mathcal{B}= \{ B_0, B_1, \dots, B_m\}} \frac{1}{2 \lvert \mathcal{P}_{\bz} \rvert} \sum_{i=1}^{m} \sum_{\bz_i \sim B_i} \lVert \bz_i -  \bc_j \rVert^2 \\
&\argmin_{\mathcal{B}= \{ B_0, B_1, \dots, B_m\}}  \frac{1}{2 \lvert \mathcal{P}_{\bz} \rvert} \sum_{i=1}^{m} | B_i | \cdot Var(B_i)
\end{align}
\vspace{-0.15in}

Here we see that minimizing the commitment loss is equivalent to computing a set of balls $\mathcal{B}$ with the mean of each ball being $\mu(B_i) = \bc_i$ such that the sum of the weight variance of all the balls is minimized. Solving this assignment problem NP-hard and cannot be trivially attained. Note that all existing algorithm for solving K-means is a heuristics. The above objective solves a global optimization assignment problem while optimizing with SGD is a greedy local update. With abuse of notation, define $B(\bc_j) = \{ \forall \; \bz_i \in \mathcal{P}_{\bz} \; \text{where} \; \bc_j = \argmin(\lVert \bz_i - \bc_j \rVert^2)\}$ to be the set of points that is closest to the center $\bc_j$. Then the commitment loss we optimize in practice is:

\vspace{-0.15in}
\begin{align}
&\min_{\bc_j} \frac{1}{2 \lvert \mathcal{P}_{\bz} \rvert} \sum_{i=1}^{m} \sum_{\bz_i \sim B(\bc_j)} \lVert \bz_i -  \bc_j \rVert^2 
\end{align}
\vspace{-0.15in}

With the update rule for the code vectors being:

\vspace{-0.15in}
\begin{align}
\bc_j^{(t+1)} \leftarrow \bc_j^{(t)} - \eta \sum_{\bz_i \in B(\bc_j)} (\bz_i - \bc_j)
\end{align}
\vspace{-0.15in}

The update rule states that each code vector moves toward the mean of its current ball. Note that the cardinality of the ball can be zero; in such case, there is no gradient for the code-vector. Hence, in the worst case where $B(\bc_j) = \emptyset$ for all $j\neq i$ for some $i$, we achieve the upper bound. Once the ball becomes empty, the code-vectors do not receive any gradients. To achieve the lower bound, one must hope that there is good coverage over $\mathcal{P}_{\bz}$, and by minimizing the loss, we recover the optimal assignment. This highlights the difficulty of achieving the optimal assignment using the commitment loss.

\subsection{Alternating optimization ablation}
\label{app:alt-opt}
In~\tbl{table:em}, we compare how changing the number of inner and outer optimization steps affects model performance. We find that the inner step of $\times8$ works the best. Increasing it further results in a negligible gain in performance.
\begin{table}
\begin{center}
\scalebox{0.65}{
\begin{tabular}{c|cccc} 
    \toprule
     & Inner. step & Outer. step & Accuracy & $\beta$ \\
    \midrule
    {\multirow{7}{*}{\rotatebox[origin=c]{90}{$\mathsf{AlexNet}$}}}
    & \multicolumn{2}{c}{$\mathsf{joint\;(baseline)}$} & 46.94 (+0.0) & 0.9    \\
    \cmidrule(lr){2-5}
    & 1 & 1                           & 52.54 (+5.58)  & 0.9  \\
    & 2 & 1                           & 55.13 (+8.19)  & 0.9   \\
    & 4 & 1                           & 57.45 (+10.51) & 0.95  \\
    & 8 & 1           &\textbf{58.03 (+11.09)} & 0.98 \\
    & 2 & 2                           & 55.09 (+8.15)  & 0.95   \\
    & 4 & 4                           & 55.29 (+8.35)  & 0.95 \\
    \bottomrule
\end{tabular}
}
\caption{\small \textbf{Alternating optimization ablation:} We vary the number of the inner and outer loops for alternated training. All models observe the same number of training examples by splitting the mini-batch.}
\label{table:em}
\end{center}
\end{table}

\subsection{Affine parameterization using EMA}
\label{app:aff-ema}
When accumulating batch statistics for affine parameters, we compute an exponential moving average over the mean and variance of $\bz_e$ and $\bz_q$ with momentum $m$:

\vspace{-0.15in}
\begin{align}
\mu_{\mathsf{ema}}(\bz_e) \leftarrow m \cdot \mu(\bz_e) + (1 - m) \cdot \mu_{\mathsf{ema}}(\bz_e) \\
\sigma^2_{\mathsf{ema}}(\bz_e) \leftarrow m \cdot \sigma^2(\bz_e) + (1 - m) \cdot \sigma^2_{\mathsf{ema}}(\bz_e) \\
\mu_{\mathsf{ema}}(\bz_q) \leftarrow m \cdot \mu(\bz_q) + (1 - m) \cdot \mu_{\mathsf{ema}}(\bz_q) \\
\sigma^2_{\mathsf{ema}}(\bz_q) \leftarrow m \cdot \sigma^2(\bz_q) + (1 - m) \cdot \sigma^2_{\mathsf{ema}}(\bz_q) \\
\end{align}
\vspace{-0.15in}

We then use these statistics to normalize the $\bz_q$. We first center the code-vectors and then re-normalize them back to the embedding moments:

\vspace{-0.15in}
\begin{align}
\bz_q \leftarrow \frac{\sigma_{\mathsf{ema}}(\bz_e)}{\sigma_{\mathsf{ema}}(\bz_q)} (\bz_q - \mu_{\mathsf{ema}}(\bz_q)) + \sigma_{\mathsf{ema}}(\bz_e)
\end{align}
\vspace{-0.15in}

The affine parameters then correspond to:

\vspace{-0.15in}
\begin{align}
\bc_{\mathsf{mean}} &= \frac{\sigma_{\mathsf{ema}}(\bz_e)}{\sigma_{\mathsf{ema}}(\bz_q} \\
\bc_{\mathsf{std}} &= \sigma_{\mathsf{ema}}(\bz_e) - \frac{\sigma_{\mathsf{ema}}(\bz_e)}{\sigma_{\mathsf{ema}}} \mu_{\mathsf{ema}}(\bz_q)
\end{align}
\vspace{-0.15in}

The momentum $m$ acts as the learning rate for the affine parameters. We find $m \in [0.01, 0.1]$ to be a good starting point for most models.

\end{document}